\pdfoutput=1
\documentclass[11pt]{article}

\usepackage[final]{acl}

\usepackage{times}
\usepackage{latexsym}
\usepackage{bbding}
\usepackage[T1]{fontenc}
\usepackage[utf8]{inputenc}

\usepackage{microtype}
\usepackage{amssymb}
\usepackage{inconsolata}

\usepackage{tcolorbox}
\usepackage{graphicx}
\usepackage{amsmath}
\usepackage{enumitem}
\usepackage{booktabs}
\usepackage{multicol,multirow}
\usepackage{adjustbox}
\usepackage{makecell}
\usepackage{longtable}

\usepackage{marvosym}

\usepackage{color, colortbl}
\usepackage{xcolor}
\definecolor{LightCyan}{rgb}{0.8, 0.9, 1}

\tcbuselibrary{skins}
\newcommand{\ourmethod}{\textsc{PUGC}}

\title{Aligning Large Language Models with Implicit Preferences from User-Generated Content}

\author{
Zhaoxuan Tan$^{\text{\Letter}\clubsuit}\thanks{\hspace{0.175cm}This work was done when Zhaoxuan and Zhihan were interns at Amazon.}$, Zheng Li$^{\diamondsuit}$, Tianyi Liu$^{\diamondsuit}$, Haodong Wang$^{\diamondsuit}$, Hyokun Yun$^{\diamondsuit}$, \\\textbf{Ming Zeng$^{\diamondsuit}$, Pei Chen$^{\diamondsuit}$, Zhihan Zhang$^{\clubsuit *}$, Yifan Gao$^{\diamondsuit}$, Ruijie Wang$^{\diamondsuit}$, }\\ \textbf{Priyanka Nigam$^{\diamondsuit}$, Bing Yin$^{\diamondsuit}$, Meng Jiang$^{\clubsuit\diamondsuit}$}\\
$^{\clubsuit}$University of Notre Dame, $^{\diamondsuit}$Amazon.com Inc \\
\texttt{ztan3@nd.edu}
}

\newtcolorbox{prompt}[1]{
    enhanced,
    drop shadow=black!5!white,
    left=4mm,
    right=4mm,
    top=2mm,
    bottom=2mm,
    boxsep=0mm,
    rounded corners,
    title=#1,    fontupper=\footnotesize\linespread{0.9}\fontfamily{lmr}\selectfont,
    }

\begin{document}
\maketitle
\begin{abstract}
% Learning from preference feedback plays a crucial role in aligning large language models (LLMs) with human values and improving generation quality. Existing preference learning methods rely heavily on preference data curated by human or strong LLM, such preference data is expensive and difficult to scale. In this work, we introduce a novel framework to harness implicit human \textbf{P}references in unlabeled \textbf{U}ser-\textbf{G}enerated \textbf{C}ontent (\ourmethod{}) for preference data generation. While UGC is not explicitly designed to guide LLMs in generating human-preferred responses, it often reflects the sharers' knowledge and insights, which can contain valuable implicit preferences. Our method transforms UGC into user queries and generates responses from the policy model. The UGC is then leveraged as a reference text for response scoring, aligning the model with the implicit preferences found in UGC. \ourmethod{} enhances the quality of preference data and allows the use of large-scale, domain-specific UGC for more adaptable alignment. Experimental results on Alpaca Eval 2 show that models trained with DPO, equipped with \ourmethod{}, achieve a 9.37\% performance improvement over traditional preference data, establishing a 35.93\% state-of-the-art length-controlled win rate using Mistral-7B-Instruct. Extensive studies also highlight improvements in reward quality, the model's theory of mind capabilities, and the effectiveness of domain-specific alignment achieved by \ourmethod{}.
Learning from preference feedback is essential for aligning large language models (LLMs) with human values and improving the quality of generated responses. However, existing preference learning methods rely heavily on curated data from humans or advanced LLMs, which is costly and difficult to scale. In this work, we present \ourmethod{}, a novel framework that leverages implicit human \textbf{P}references in unlabeled \textbf{U}ser-\textbf{G}enerated \textbf{C}ontent (UGC) to generate preference data. Although UGC is not explicitly created to guide LLMs in generating human-preferred responses, it often reflects valuable insights and implicit preferences from its creators that has the potential to address readers' questions. \ourmethod{} transforms UGC into user queries and generates responses from the policy model. The UGC is then leveraged as a reference text for response scoring, aligning the model with these implicit preferences. This approach improves the quality of preference data while enabling scalable, domain-specific alignment. Experimental results on Alpaca Eval 2 show that models trained with DPO and \ourmethod{} achieve a 9.37\% performance improvement over traditional methods, setting a 35.93\% state-of-the-art length-controlled win rate using Mistral-7B-Instruct. Further studies highlight gains in reward quality, domain-specific alignment effectiveness, robustness against UGC quality, and theory of mind capabilities. Our code and dataset are available at \url{https://zhaoxuan.info/PUGC.github.io/}.
\end{abstract}

\section{Introduction}
% LLM shows emergent ability, ..., instruction following and respond, these capabilities needs to align LLMs' response with human preferencens. 
% Large language models (LLMs) are advancing the ability to follow instructions and respond to user queries \cite{wei2021finetuned, dubey2024llama,chung2024scaling}. An important step is to align LLMs with human values and intentions, ensuring they are helpful, honest, and harmless \cite{askell2021general}. Reinforcement learning from human feedback (RLHF) \cite{ouyang2022training} is a promising approach to better align LLMs with human values. It uses a reward model to model human preferences, and steers the policy LM to maximize this reward model for desired response \cite{schulman2017proximal}. While the classical RLHF approach \cite{ouyang2022training} achieves impressive results, yet it presents optimization challenges due to its multi-stage procedure and make the training process engineering extensive and unstable \cite{casper2023open}. 
Advancements in large language models (LLMs) have enhanced their ability to follow instructions and respond to user queries \cite{wei2021finetuned, dubey2024llama, chung2024scaling}, primarily through alignment with human values to ensure they are helpful, honest, and harmless \cite{askell2021general}. Reinforcement learning from human feedback (RLHF) \cite{ouyang2022training} achieves this alignment by using a reward model that captures human preferences to guide policy models \cite{schulman2017proximal}, but its multi-stage process is complex, resource-intensive, and prone to instability \cite{casper2023open}. Direct Preference Optimization (DPO) \cite{rafailov2024direct} simplifies alignment by directly training policy models from preference data without an explicit reward model, leading to widespread adoption due to its simplicity and stability \cite{meng2024simpo}. However, both RLHF and DPO rely on high-quality preference data, which is expensive and hard to scale as it typically requires human annotators \cite{kopf2024openassistant} or stronger models like GPT-4 \cite{cui2023ultrafeedback}, and inconsistent data quality further limits alignment performance \cite{zhou2024lima, wu2024meta, liu2023makes, zheng2023secrets}.

% Large language models (LLMs) are continually improving in their ability to follow instructions and respond to user queries \cite{wei2021finetuned, dubey2024llama, chung2024scaling}. A critical aspect of this progress is aligning LLMs with human values and intentions to ensure they are helpful, honest, and harmless \cite{askell2021general}. Reinforcement learning from human feedback (RLHF) \cite{ouyang2022training} is a promising method for achieving this alignment. RLHF works by using a reward model that captures human preferences and directs the policy model to maximize this reward, thereby producing desired responses \cite{schulman2017proximal}. While RLHF has demonstrated impressive results, its multi-stage procedure introduces optimization challenges, making the training process complex, resource-intensive, and prone to instability \cite{casper2023open}.

\begin{figure}[t]
    \centering
    \includegraphics[width=1\linewidth]{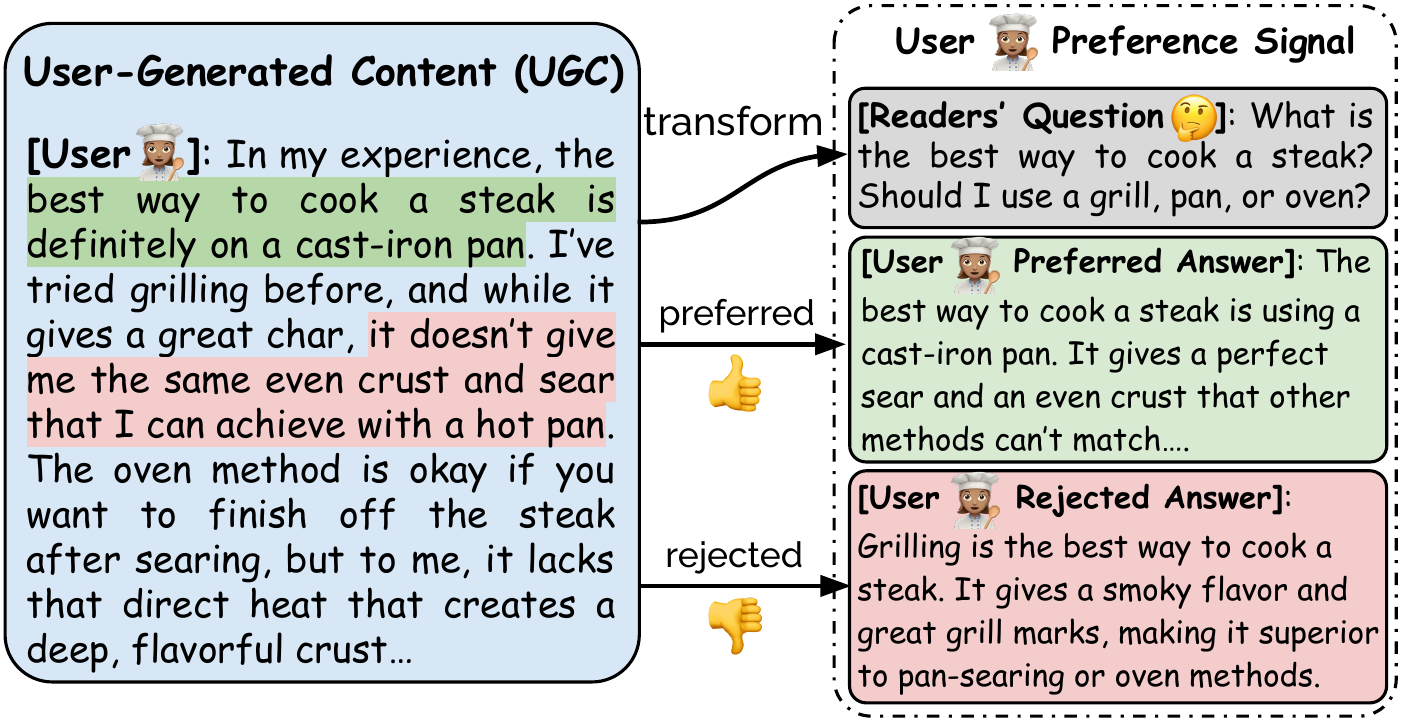}
    \caption{User-Generated Content (UGC) is created to share users' knowledge and insights with others, often addressing readers' questions. This creates the potential to transform UGC into reader queries and sharer-preferred answers, which can be utilized to align LLMs with user preferences.}
    \label{fig:teaser}
\end{figure}

Among the potential solutions to align LLMs with human preference without explicit human and powerful model annotation, leveraging large amounts of unlabeled data to align LLMs with human preference emerges as a promising approach. Existing methods focus on sourcing supervised fine-tuning (SFT) data from unlabeled text. HumpBack \cite{li2023self} translates web corpus data into instructions and fine-tunes models using high-quality, self-curated instructional data. Back-and-forth translation \cite{nguyen2024better} takes this further by applying additional filtering and rewriting steps to enhance instruction and response quality. However, research shows that SFT provides a biased estimate of human preferences by relying on predefined input-output pairs. In contrast, RLHF helps further align the model with human preferences by dynamically updating its responses based on feedback \cite{kirk2023understanding, dong2024rlhf}. The research question arises: \textit{Can we collect preference data for LLM response distantly from the massive unlabeled data?}

% Another line of work focus on using the unlabeled data to align LLMs using instruction backtranslation, and back and forth translation to conduct instruction finetuning. However, the human curated dataset can be expensive to create and the quality is hard to control and limited the LLM performance. and the instruction tuning shows limited adapatability to preferences and less flexibility in generalization and requires high-quality instructional data.

%  Among the massive text corpus, the text sources such as tweets, reviews, and blogs, that reflect the implicit user preferences collectively referred to as user-generated content (UGC) can be a good source of user preference to align LLMs with human preferences. UGC is abundant on the internet through platforms like Amazon reviews, Reddit, and StackExchange posts, and we aim to utilize this vast resource to help LLMs generate responses that better align with human preferences. Although UGC is not explicitly created to annotate user preferences, it often reflects the sharer's knowledge and insights, providing potential answers to readers' questions. These insights can be transformed into reader questions and reflect the sharer's preferences in answers (Figure \ref{fig:teaser}). 
% By transforming UGC into readers' queries and sharers' preferences, we can extract valuable supervision for human preference alignment, constructing high-quality and large-scale preference data to improve LLMs. 

Among the vast amounts of textual data available, user-generated content (UGC)—including tweets, reviews, and blogs that contain user's insight and knowledge—serves as a valuable source for aligning LLMs with human preferences. UGC is abundant on platforms like Amazon reviews, Reddit, and StackExchange, making it a scalable resource. Although UGC is not explicitly created to annotate user preferences, it often contains the sharer's knowledge and insights, providing potential answers to readers' questions. These insights can be transformed into reader queries, capturing the sharer's preferences in their responses (Figure \ref{fig:teaser}). By converting UGC into reader queries and sharer preferences, we can extract valuable supervision for human preference alignment, thereby constructing high-quality, large-scale preference data to improve LLMs.

Leveraging the implicit preferences in UGC, we propose \ourmethod{}, a novel framework to generate high-quality preference data from UGC for aligning LLMs. First, \ourmethod{} employs an LLM to extract potential reader queries from UGC. We then use the same LLM to verify that the UGC contains sufficient information relevant to the generated instructions, filtering out irrelevant content. Next, we feed these instructions into a policy model and sample multiple responses. \ourmethod{} evaluates these responses using a reward model, with the original UGC serving as a reference to uncover implicit user preferences. Thus, \ourmethod{} efficiently leverages unlabeled UGC for auxiliary supervision, enhancing reward quality compared to traditional preference data generation paradigms and enabling more flexible alignment across domains without collecting human instructions.

% \ourmethod{} leverages unlabeled UGC for auxiliary supervision, enabling distant human guidance to enhance preference data. Additionally, \ourmethod{} avoids the need for collection of real-world, domain-specific instruction data, making domain-specific alignment more flexible.

Our experiments show that using 60k UGC from Dolma dataset \cite{soldaini-etal-2024-dolma}, \ourmethod{} can obtain a model that achieves the state-of-the-art length-controlled win rate of 35.93\% based on Mistral-7B-Instruct-v0.2 on AlpacaEval 2.0 benchmark. \ourmethod{} supports different preference-tuning methods such as DPO \cite{rafailov2024direct} and SimPO \cite{meng2024simpo}, and showing consistent superiority compared to preference data derived from UltraFeedback \cite{cui2023ultrafeedback} instructions. The performance advantage is consistent across various training setups and diverse instruction-following benchmarks, including AlpacaEval 2 \cite{alpaca_eval} and MT-Bench \cite{zheng2023judging}. We also observe a significant performance gain in the theory of mind capabilities demonstrated by BigGen Bench \cite{kim2024biggen}. Further studies also reveal \ourmethod{}'s improved reward quality, robustness against UGC quality, comprehensive performance gain in fine-grained Alpaca Eval analysis, and its effectiveness and flexibility in domain-specific alignment.

% To summarize, \ourmethod{} is the first to harnesses implicit user preferences from UGC to construct high-quality preference data in a cost-efficient manner. This approach enables flexible, domain-specific alignment and demonstrates significant advantages over traditional preference data generation methods. By introducing this novel pipeline, \ourmethod{} opens new avenues for utilizing unlabeled text in human preference alignment.

In summary, \ourmethod{} is the first to harness implicit user preferences from UGC to cost-effectively construct high-quality preference data. This approach enables flexible, domain-specific alignment and offers significant advantages over traditional preference data generation methods. By introducing this novel pipeline, \ourmethod{} opens new avenues for utilizing unlabeled text in human preference alignment.

\section{Preference Alignment using UGC (\ourmethod{})}
In this section, we first provide the background of current preference data generation pipeline in \S \ref{subsec:background}. We then introduce our proposed \ourmethod{} pipeline (\S \ref{sec:method}), which leverages implicit user preference signals from UGC, highlighting how it differs from traditional preference data generation approach.

\subsection{Traditional Preference Data Generation}
\label{subsec:background}
In traditional preference data generation pipelines for alignment, a set of prompts $\mathcal{X}$ is collected, and multiple responses are sampled for each prompt $x\in \mathcal{X}$ from the policy model $\pi_\theta$. 
\begin{align*}
    \{a_{i}\}_{i=1}^{K} \sim \pi_\theta (a \mid x),
\end{align*}
where $a$ denotes the generated response and $K$ denotes the number of generated responses for each prompt. Once the responses are sampled, a reward model $\textit{RM}$ scores each response $a_i$ for prompt $x$:
\begin{align*}
    y_i = \textit{RM}(a_i \mid x ),
\end{align*}
where $y_i$ is the reward score corresponding to response $a_i$. After scoring all the responses $\{a_{i}\}_{i=1}^{K}$, the responses with the highest and lowest scores are selected as the preferred response $a_w$ and rejected response $a_l$, respectively. This forms the preference data pair $(x, a_{w} > a_{l})$ which is used for preference tuning.

\begin{figure*}[t]
    \centering
    \includegraphics[width=0.9\linewidth]{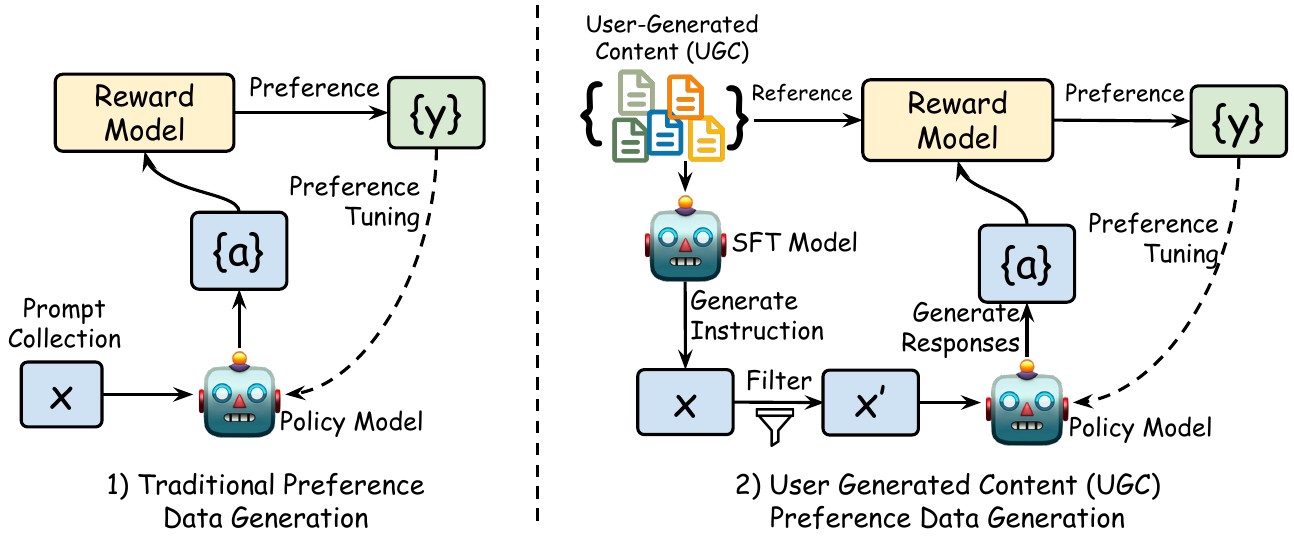}
    \caption{Overview of \ourmethod{}. Unlike traditional preference data generation methods, \ourmethod{} sources preference data from user-generated content (UGC) by transforming UGC into reader questions and using UGC as a reference for preferred answers. \ourmethod{} leverages implicit user preference signals to align LLMs.}
    \label{fig:overview}
\end{figure*}

\subsection{Generating Preference Data from UGC}
\label{sec:method}

Motivated by the fact that UGC is often created to address potential questions from other users, \ourmethod{} begins by transforming UGC into a reader's question using an LLM (SFT version model by default). This transformation can be expressed as:
\begin{align*}
    x = \textit{GenInst}((\phi_q(d)) \mid \pi_\theta),
\end{align*}
% where $\phi_q$ represents the prompt generation template. We retain only the instructions for which the corresponding $s_x$ is $\textit{True}$ to ensure instruction quality. These filtered instructions are denoted as $\overline{x}$. To leverage the sharers' insights, knowledge, and implicit preference signals within the UGC, we then use the UGC as reference text in the reward model for response scoring, mining implicit preference signals from the UGC. The reward score for the $i$-th response is given by:
where $\phi_q$ denotes the prompt construction function that converts the UGC $d$ into a reader instruction $x$. To ensure the quality of the generated instructions, we filter out those that are irrelevant to the source UGC. Specifically, we utilize the same LLM for instruction generation $\pi_\theta$ to assess whether the UGC contains sufficient information to answer the generated instruction. This evaluation produces a filtering score $s \in \{\textit{True}, \textit{False}\}$ for each instruction-UGC pair.
\begin{align*}
     s_x = \textit{Filter}(\phi_f(x, d) \mid \pi_\theta),
\end{align*}
where $\phi_f$ is the filtering prompt template. We then only keep the instruction and UGC that corresponding $s_x$ is $\textit{True}$ to ensure the instruction quality, the kept instructions are denoted as $x'\in \mathcal{X}'$. To leverage the sharers' insights and knowledge and implicit preference signals within the UGC, we then directly feed the UGC as a reference answer in the reward model for response scoring, mining the implicit preference signals in the UGC, where the reward score for the $j$-the response is
\begin{align*}
    y_j = \textit{RM}(a_i \mid x', d).
\end{align*}
Finally, using the instruction $x'$, the sampled responses $\{a_j\}$, and the corresponding reward scores $\{y_j\}$, we can obtain the preference data pairs $\mathcal{P}$ for the entire instruction set $\mathcal{P} = \{(\overline{x}, y_w > y_l)\}, \overline{x}\in{\mathcal{X}'},$
where $y^w$ is the chosen response, with $w = \arg\max_j(\{y_j\}_{j=1}^{K})$, and $y_l$ is the rejected response, with $l = \arg\min_j(\{y_j\}_{j=1}^{K})$. 

\ourmethod{} is a versatile preference data generation framework that leverages implicit user preference signals from unlabeled UGC, requiring only UGC instead of prompt instructions. Since domain-specific UGC is easier to collect compared to user instructions, \ourmethod{} offers a more flexible framework for domain-specific alignment. Moreover, \ourmethod{} is compatible with various preference tuning methods, such as DPO \cite{rafailov2024direct}, SimPO \cite{meng2024simpo}, and others.

% \ourmethod{} is a versatile preference data generation framework that leverage the implicit user preference signals in the unlabeled UGC, requesting the unlabeled UGC instead of prompt instruction. Since domain-specific UGC is easier to collect compared to the user instruction, \ourmethod{} is also a more flexible framework for domain-specific alignment. Moreover, \ourmethod{} is a versatile framework that compatible with diverse preference tuning methods that request preference data pairs, such as DPO \cite{rafailov2024direct}, IPO \cite{azar2024general}, SimPO \cite{meng2024simpo}, and more.

\section{Experiment Settings}

\paragraph{Datasets} For the UGC data, we select the Dolma \cite{soldaini-etal-2024-dolma} pretraining corpus. Specifically, we sample data from \textit{cc\_en\_head}, \textit{megawika}, \textit{stackexchange}, \textit{tulu\_flan}, \textit{open\_web\_math}, and \textit{starcoder} as the text corpus, and use Llama-3-70B-instruct \cite{dubey2024llama} via Bedrock API to generate a 5-point quality score for the unlabeled text\footnote{Appendix \ref{sec:ugc_quality_selection} and \ref{sec:ugc_quality_validation} justify model selection and validate scoring quality.}. To balance data quantity and quality, we select UGC with scores greater than or equal to 4 and feed them into the UGC preference data generation pipeline.

\paragraph{Models and Training Settings} We perform preference tuning with Mistral-7B \cite{jiang2023mistral} model under the base and instruct setup following \citet{meng2024simpo}. For the base setup, we use \texttt{zephyr-7b-sft-full} \cite{tunstall2023zephyr}, trained on the UltraChat-200k \cite{ding2023enhancing}, which provides a high level of transparency. For the instruct setup, we use the off-the-shelf instruction-tuned model \texttt{Mistral-7B-Instruct-v0.2}, which is more robust but less transparent as the RLHF process is not publicly disclosed. For response sampling from the policy model, we follow the setting in SimPO and set the temperature to 0.8 and use nucleus sampling with $p$=0.95. The reward model used is \texttt{prometheus-7b-v2.0} \cite{kim2024prometheus}, chosen for its strong judgment correlation with GPT and human, and the ability to directly score responses using reference answers. We adopt self-consistency decoding with $N$=8 for reward score generation, as suggested by \citet{kim2024biggen}, and resolve ties by selecting the shortest high-scoring response as the preferred answer and the longest low-scoring response as the rejected answer to mitigate length bias.

% For the base setup, we adopt the \texttt{alignment-handbook/zephyr-7b-sft-full} \cite{tunstall2023zephyr}, which is trained on the UltraChat-200k \cite{ding2023enhancing} dataset to obtain an SFT model. The base setup provides a high level of transparency. For the Instruct setup, we use the off-the-shelf instruction-tuned model (\texttt{mistralai/Mistral-7B-Instruct-v0.2}) as the SFT model, which is more powerful and robust than the base SFT model, but more opaque since the RLHF procedure is not publicly disclosed. For the reward model, we select the \texttt{prometheus-eval/prometheus-7b-v2.0} \cite{kim2024prometheus} since it naturally accept the reference answer in direct scoring mode and shows stronger generalization across different domains. We adopted self-consistency decoding and set $N$=5 as is suggested by \citet{kim2024biggen}. To break the tie, we select the shortest answer with the highest score as chosen response and longest answer with the lowest score as the rejected response to mitigate the length bias in preference tuning.

\paragraph{Evaluation Benchmark and Evaluation Metrics} We evaluate our method on two widely used open-ended generation quality benchmarks: MT-Bench \cite{zheng2023judging} and AlpacaEval 2 \cite{alpaca_eval}. AlpacaEval 2 includes 805 questions from five datasets, using GPT-4 Turbo as a judge for pairwise comparisons between the evaluated model's generation and GPT-4 Turbo's output. We report both raw win rate and length-controlled win rate \cite{dubois2024length} to mitigate verbosity bias. MT-Bench consists of 80 multi-turn questions across 8 categories, and we report the average MT score using GPT-4 as the judge.\footnote{Decoding hyperparameters are provided in Appendix \ref{decode_hyper} for reproducibility.}

% We primarily assess our method on two most popular open-ended generation quality benchmarks: MT-Bench \cite{zheng2023judging} and AlpacaEval 2 \cite{alpaca_eval}. AlpacaEval 2 consists of 805 questions from 5 datasets, and use GPT-4 Turbo as a judge to do pairwise comparison between evaluated model generation and GPT-4 Turbo generation. We report raw win rate and length-controlled win rate \cite{dubois2024length} to eliminate the model verbosity. MT-Bench covers 8 categories with 80 multi-turn questions, and we report the average MT score using GPT-4 as the judge model. We also report the arena-hard \cite{li2024crowdsourced} and downstream task from the huggingface OpenLLM Benchmark performance in the Appendix \ref{openllm}. Decoding hyperparameters are presented in Appendix \ref{decode_hyper} to facilitate reproducibility.

\paragraph{Baselines}
We compare \ourmethod{} against offline preference optimization methods using on-policy preference data generated via the traditional pipeline with 60k prompts from UltraFeedback \cite{cui2023ultrafeedback}, ensuring a fair comparison. Preference optimization methods include ORPO \cite{hong2024reference}, KTO \cite{ethayarajh2024kto}, DPO \cite{rafailov2024direct}, R-DPO \cite{park2024disentangling}, CPO \cite{xu2024contrastive}, and SimPO \cite{meng2024simpo}. All baselines employ the same reward model, \texttt{prometheus-7b-v2.0}, with identical settings to those of our proposed \ourmethod{}, but without reference text for reward scoring to maintain fairness in the comparison.

% We compare \ourmethod{} with offline preference optimization methods using preference data generated through the traditional pipeline with 60k prompts from UltraFeedback \cite{cui2023ultrafeedback}, ensuring a fair comparison. Preference optimization methods include ORPO \cite{hong2024reference}, KTO \cite{ethayarajh2024kto}, DPO \cite{rafailov2024direct}, R-DPO \cite{park2024disentangling}, CPO \cite{xu2024contrastive}, and SimPO \cite{meng2024simpo}. For fair comparison, the reward model adopted in baseline is the same \texttt{prometheus-eval/prometheus-7b-v2.0} follow the same setting but without reference text.

\begin{table*}[t]
\centering
    \caption{Alpaca Eval 2.0 and MT-bench results. LC and WR denotes length-controlled and raw win rate, respectively. We use the off-the-shelf model as the SFT model. We use GPT-4-Turbo as a judge for MT-Bench scoring. The best score is in \textbf{bold} and the second best is \underline{underlined}.}
    \begin{adjustbox}{max width=0.9\linewidth}
    \begin{tabular}{lcccccccc}
    \toprule
     \multirow{3}{*}{\textbf{Method}}& \multicolumn{4}{c}{\textbf{Mistral-Base (7B)}}& \multicolumn{4}{c}{\textbf{Mistral-Instruct (7B)}}\\
     \cmidrule(lr){2-5}\cmidrule{6-9}
     &\multicolumn{3}{c}{\textbf{Alpaca Eval 2.0}}& \multicolumn{1}{c}{\textbf{MT-Bench}}& \multicolumn{3}{c}{\textbf{Alpaca Eval 2.0}}& \multicolumn{1}{c}{\textbf{MT-Bench}} \\
     \cmidrule(lr){2-4}\cmidrule(lr){5-5}\cmidrule(lr){6-8}\cmidrule{9-9}
     &LC (\%) & WR (\%)&  Len. & GPT-4-Turbo & LC (\%) & WR (\%) &  Len. & GPT-4-Turbo \\
     \midrule
     \textsc{SFT}& 7.32 & 4.41 & 928 & 4.68 &  17.11 & 14.72 & 1676 & 6.20 \\
     \midrule
     \multicolumn{9}{l}{\ \ \textit{On-Policy Preference Data Generated with \textsc{UltraFeedback}}}\\
    \textsc{ORPO}& 8.18 & 4.66 & 867 & 5.04 & 20.85 & 16.75 & 1574 &  6.18 \\
     \textsc{KTO}& 7.44 & 6.31 & 1367 & 4.92 & 22.45 & 20.07 & 1741 & 6.04\\
     \textsc{DPO}& 15.60 & \underline{10.23} & 1118 & 5.04 & 26.56 & 24.39 &1750 & 6.04\\
     % IPO& 17.63 & 12.42 & 1151 & & 29.75 & 34.95 & 2427\\
     \textsc{R-DPO}& 15.23 & 8.72 & 922 & \textbf{5.14} & 27.46 & 23.75 & 1658 & 6.06\\
     \textsc{CPO} & 9.17 & 5.50 & 932 & 4.75 & 21.34 & 19.40 & 1802 & \underline{6.22}\\
     \textsc{SimPO}& 14.84 & 8.59 & 733 & 4.71 & 29.16 & \underline{32.92}  & 2338 & 5.66\\
     \midrule
     \multicolumn{9}{l}{\ \ \textit{Preference Data Generated with \ourmethod{} (Ours)}}\\
     % \multicolumn{9}{l}{\textit{Instruction generated from \textbf{UGC}}\\
     % & Snorkel-Mistral & 26.4\% & 30.2\% & 2736 & 7.6 \\
     % \cmidrule(lr){2-10}
     \textsc{\ourmethod{}+DPO} & \textbf{17.57} & \textbf{11.30} & 1013 & \underline{5.13} & \textbf{35.93} & \textbf{33.05} & 1705 & \textbf{6.27} \\     \textsc{\ourmethod{}+SimPO} & \underline{16.95} & 9.10 & 700 & 5.09 & \underline{32.15} & 30.34 & 1844 & 6.13\\
     % \midrule
     % \multicolumn{9}{l}{\ \ \textit{Claude-3-Sonnet Generated Instructions from \textsc{UGC}}}\\

     % \textsc{\ourmethod{}+DPO} & 17.57 & 11.30 & 1013 & 5.13 & 35.71 & 34.00 & 1845 & 6.27\\
     % \textsc{\ourmethod{}+SimPO} & 16.43 & 9.85 & 793 & & 31.52 & 28.82 & 1792 & 5.83\\
     \bottomrule
    \end{tabular}
    \end{adjustbox}
    \label{tab:main_result}
\end{table*}

\section{Results}
Table \ref{tab:main_result} shows the performance of \ourmethod{} and baselines on Alpaca Eval 2.0 and MT-Bench, we have the following observations.
\paragraph{\ourmethod{} significantly outperform baseline}
Models trained with \ourmethod{} outperform baseline methods that use UltraFeedback preference data. Notably, when using the DPO objective, \ourmethod{} achieves an absolute improvement in LC win rate of 2.0\% and 9.4\% on the Alpaca Eval 2.0 benchmark under the base and instruct settings, respectively. Similarly, models trained with SimPO show improvements of 2.1\% and 3.0\% in both settings. On MT-Bench, \ourmethod{} also demonstrates superiority over models trained with UltraFeedback data and the SFT baseline. The limited performance gains or occasional degradation could be attributed to the limited judging capability of the Prometheus-2-7B model in reasoning, coding, and math domains, and the limited UGC in related domains. Additionally, the small evaluation dataset and single-instance assessments may impact the quality of MT-Bench results.
% Moreover, \ourmethod{} mainly focuses on improving the response quality and insightfulness with UGC, while the there lacks such reasoning data in UGC data, potentially lead to the limited performance gain in reasoning capability.

\paragraph{\ourmethod{} shows moderate length exploitation}
While \ourmethod{} achieves significant improvements, moderate length exploitation is observed in both the base and instruct setups. Models trained with SimPO and DPO generate responses that are relatively shorter than those from the baselines, yet they still achieve higher win rates. This suggests that \ourmethod{}'s preference data generation pipeline can better mitigate length bias.

\paragraph{\ourmethod{}+DPO generally shows the best performance}
% Despite much variant of DPO has been proposed, the performance of model trained with different objectives under our setting shows that DPO generally perform the best. 
% Despite the existence of various preference optimization objectives, DPO consistently delivers the best results when trained with \ourmethod{} and UltraFeedback preference data. For example, \ourmethod{}+DPO achieves a length-controlled win rate of 35.93\% in the Mistral-Instruct setup, outperforming other methods, including SimPO. This suggests that DPO, when combined with \ourmethod{}'s preference data generation, remains a highly effective optimization strategy for aligning LLMs with human preferences.
Trained on UltraFeedback data, DPO and SimPO generally perform the best across both settings, which justifies the choice of applying DPO and SimPO to \ourmethod{} preference data. Although SimPO outperforms DPO in the Instruct setting, \ourmethod{}+DPO achieves a higher LC win rate compared to \ourmethod{}+SimPO in both settings. This indicates that DPO, when paired with \ourmethod{} preference data, remains a highly effective strategy for aligning LLMs with human preferences.

% \vspace{-0.25cm}

\paragraph{The Instruct setting introduces significant performance gains} Across both Alpaca Eval 2.0 and MT-Bench, we observe that Instruct setting consistently outperforms the Base setting. This improvement is likely due to the higher quality of SFT model used for instruction generation, as well as for preference tuning initialization and the generation of higher-quality preference by the Instruct setting models.

\begin{figure}
    \centering
    \includegraphics[width=1\linewidth]{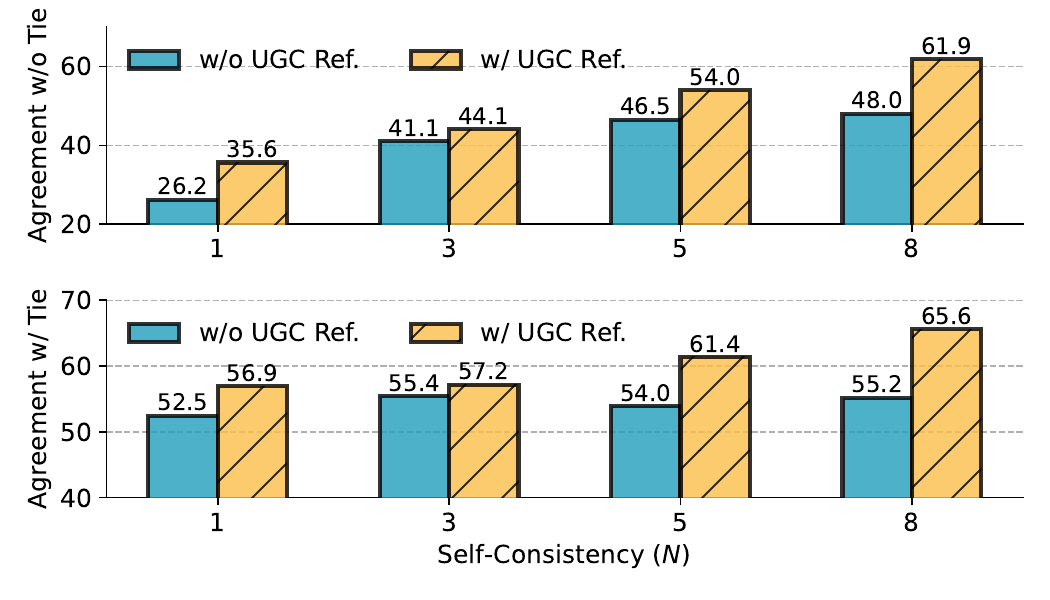}
    \caption{Reward judge agreement with GPT-4-Turbo on \ourmethod{}-sampled responses, under varying self-consistency settings. Reference UGC improves reward agreement, with higher self-consistency $N$ leading to greater improvements.}
    \label{fig:reward_quality}
    \vspace{-0.5cm}
\end{figure}

\section{Analysis}
% \vspace{-0.3cm}

\paragraph{UGC Reference Improves Reward Quality}
\label{sec:reward_quality}
To validate the reward quality in \ourmethod{} generated data and the impact of UGC preference signals, we evaluate the judging accuracy on responses sampled from model-generated instructions and alter the number of $N$ in self-consistency decoding. Specifically, we generate two responses using the policy model for each prompt, then generate preference labels with GPT-4-Turbo judge using the prompt in Alpaca Eval. To mitigate the position bias, we make two judgments with switched positions for each response pair and retain the data where two judgments agree. This process yields a total of 202 pairs with preference judge labels. Subsequently, we use the reward model to generate reward score for those responses, under both with reference UGC and without UGC reference settings. We calculate two metrics: agreement (counting ties as 0.5) and agreement without ties (ignore ties). As is shown in Figure \ref{fig:reward_quality}, judging responses with UGC as reference yields average 8.45\% and 6\% absolute improvement under with tie and without settings. Moreover, by increasing the number of sampling $N$ in self-consistency decoding, the reward quality improves, while higher $N$ can lead to greater improvements. These results demonstrate the effectiveness of the introducing reference UGC in harnessing the implicit preference signals, bringing its evaluations substantially closer to those of more sophisticated language models like GPT-4-Turbo.

\begin{figure}[t]
    \centering
    \includegraphics[width=1\linewidth]{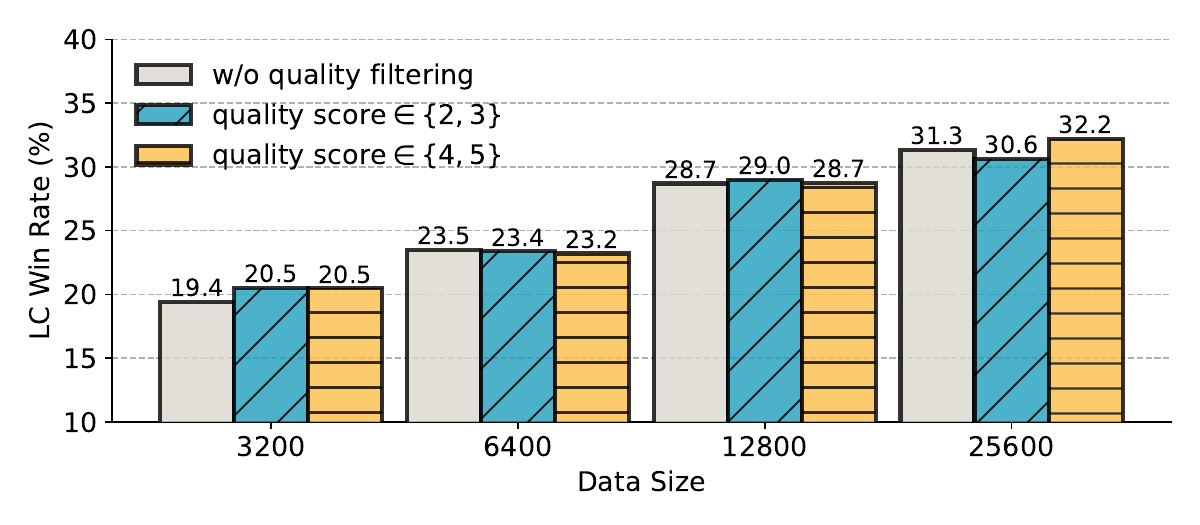}
    \caption{Alapaca Eval 2.0 performance of models trained with varying quality and quantity of preference data generated by \ourmethod{}. Increasing data quantity significantly improves performance, while \ourmethod{} remains robust to variations in UGC quality.}
    \label{fig:quality_quantity}
\end{figure}

\paragraph{UGC Quality \emph{v.s.} Quantity}

To assess the impact of UGC quality and quantity in aligning LLMs with human preferences, we compare models trained on augmented data with varying levels of quality and quantity. Specifically, we evaluate models trained on UGC without quality filtering, with quality scores in the range of ${2, 3}$, and with scores in the range of ${4, 5}$, across different data sizes. As shown in Figure \ref{fig:quality_quantity}, model performance improves significantly as data size increases, while UGC quality has a relatively minor effect on alignment performance. It demonstrates the robustness of \ourmethod{} to UGC quality variations, with larger data quantities yielding substantial performance gains.

% In order to understand the importance of UGC quality and quantity in aligning LLMs with human preferences, we compare training on augmented data of different quality and quantity. Specifically, we compare model trained with UGC without quality filtering, quality score $\in\{2, 3\}$, and quality score $\in\{4,5\}$ categories under different size. As is shown in Figure \ref{app:quality_filter}, we find that as the data size scale up, the model performance grows significantly, while the UGC doesn't have large impact on the alignment performance. These results demonstrate the proposed \ourmethod{}'s robustness against UGC quality, whereas the increased quantity can provide significant gains. 

\begin{figure}
    \centering
    \includegraphics[width=1\linewidth]{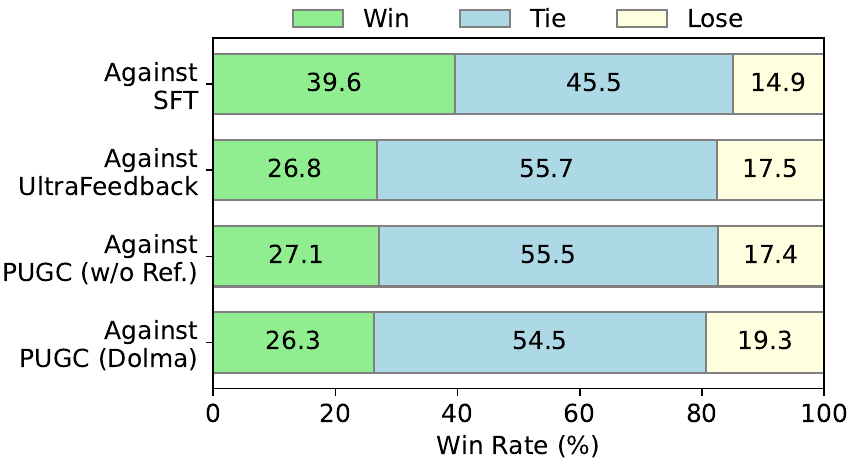}
    \caption{Domain-specific alignment performance of head-to-head win rate of \ourmethod{} trained with Goodreads book review against diverse baselines judged by GPT-4-Turbo. \ourmethod{} (Goodreads) has a 7\% higher win rate against the strongest baseline.}
    \label{fig:domain_alignment}
\end{figure}

\paragraph{Domain-Specific UGC Alignment}
\label{domain_specific_alignment}
As \ourmethod{} adapts to different domains using domain-specific UGC, we evaluate its effectiveness with user book review data from Goodreads\footnote{\url{https://www.goodreads.com/}} \cite{wan-etal-2019-fine}. For evaluation, we use the MT-Bench pairwise prompt with GPT-4-Turbo to compare \ourmethod{} (Goodreads) against the SFT baseline, the model trained on UltraFeedback preference data, and \ourmethod{} (general UGC from Dolma), and model trained on Goodreads review but without reference UGC in reward scoring. Evaluation prompts are generated by Claude-3-sonnet via Bedrock API based on book reviews to avoid distribution overlap with the SFT model. As shown in Figure \ref{fig:domain_alignment}, \ourmethod{} (Goodreads) consistently outperforms all four baselines, with a 7\% higher win rate over \ourmethod{} (Dolma). These results demonstrate that \ourmethod{} is capable of generating high-quality, domain-specific preference data for more effective and flexible alignment.

\begin{figure*}
    \centering
    \includegraphics[width=1\linewidth]{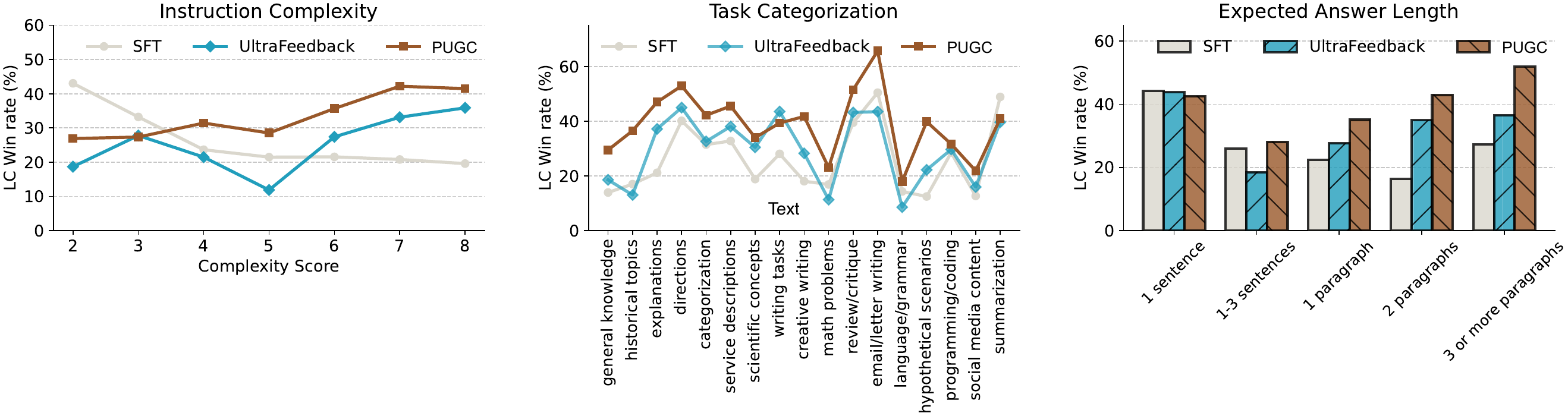}
    \caption{Fine-grained Alpaca Eval LC win rate by categorizing test instructions based on instruction complexity, task categorization, and expected answer length, showing particularly strong gains in handling more complex instructions and those requiring longer responses. We use DPO objective for preference tuning.}
    \label{fig:finegrain_alpaca}
\end{figure*}

\paragraph{Fine-Grained Alpaca Eval LC Win-Rate Analysis}
We analyze the Alpaca Eval test set to identify the task types contributing to the observed improvements by categorizing the set based on instruction complexity, task type, and expected response length. This grouping is achieved using prompts from Claude-3-Sonnet, with details and clustering statistics provided in Appendix \ref{app:alpaca_finegrained}. The results in Figure \ref{fig:finegrain_alpaca} reveal the following insights: \textit{i)} \ourmethod{} significantly improves performance on more complex instructions (complexity $\geq$ 4), with greater complexity leading to larger gains, likely due to the limited reward quality of the Prometheus model on simpler instructions.
\textit{ii)} \ourmethod{} substantially boosts LC win rates across most task categories, except for the math problems, programming, and summarization tasks, likely due to the lack of UGC in related domains and the distribution shift of instructions sourced from UGC. Compared to models trained with UltraFeedback data, \ourmethod{} shows notable improvements in categories like general knowledge, historical topics, review/letter writing, critique, and hypothetical scenarios, which are common in user-generated content. Moreover, the Prometheus-2 model was not trained on reasoning-intensive tasks, such as coding and math, which also limited the alignment performance. We leave developing a more domain comprehensive reward model that can accept the reference text as future work.
\textit{iii)} \ourmethod{} also shows significant LC win rate increases on tasks expecting outputs longer than one sentence, possibly due to length bias in UGC, as longer content tends to provide more insightful preference signals.

% \paragraph{\ourmethod{} improves answering hard and complex questions}
\begin{table}[t]
    \centering
    \caption{Theory of Mind capability evaluated on BigGen Bench \cite{kim2024biggen}, with scoring performed by GPT-4-Turbo, GPT-4, and Prometheus-bgb-8x7b-v2.0 as judge models. }
    \begin{adjustbox}{max width=1\linewidth}
    \begin{tabular}{l c c c c }
         \toprule[1.5pt]
        % \hline
        \multirow{2}{*}{\textbf{Model}} & \multicolumn{4}{c}{\textbf{BigGen Bench (ToM)}} \\
         \cmidrule(r){2-5}
         & GPT-4-Turbo & GPT-4 & Prometheus2 & Avg. \\
          \midrule[0.75pt]
         gpt-4-0125-preview & 4.26 & 4.21 & 4.15 & 4.21 \\
         \rowcolor{LightCyan}
         \ourmethod{} + DPO & 4.22 & 4.10 & 4.17 & 4.16 \\
         gpt-4-1106-preview & 4.24 & 4.07 & 4.09 & 4.13 \\
         gpt-4o-2024-05-13 & 4.08 & 4.04 & 4.06 & 4.07 \\
         gpt-4-turbo-2024-04-09 & 4.12 & 4.03 & 4.06 & 4.07 \\
         Claude-3-Opus & 4.09 & 4.08 & 4.05 & 4.07 \\
         LLama-3-70B-Instruct & 4.04 & 3.92 & 3.97 & 3.98 \\
         \rowcolor{LightCyan}
         \ourmethod{} + SimPO & 4.03 & 3.89 & 3.84 & 3.92\\
         Mixtral-8x7B-Instruct-v0.1 & 3.81 & 3.80 & 3.78 & 3.80 \\
         Meta-Llama-3-8B-Instruct & 3.83 & 3.65 & 3.71 & 3.73 \\
         Mistral-7B-Instruct-v0.2 & 3.77 & 3.68 & 3.49 & 3.65 \\
         Nous-Hermes-Mistral-7B-DPO & 3.69 & 3.51 & 3.46 & 3.55 \\
         % zephyr-7b-beta & 3.73  & 3.64  & 3.50 & 3.62 \\
         % Llama 3.1 8B Instruct & \\
        
         \bottomrule[1.5pt]
    \end{tabular}
    \end{adjustbox}
    \label{tab:tom}
    % \vspace{-0.4cm}
\end{table}

\paragraph{\ourmethod{} Significantly Improves Theory-of-Mind Capability}
As defined in BigGen Bench \cite{kim2024biggen}, the theory of mind objective evaluates whether a language model can understand another individual's beliefs, intentions, and emotions through discourse, narratives, or dialogue. User-generated content (UGC) often reflects these aspects, providing rich theory of mind knowledge that can enhance the preference data generated by \ourmethod{}. To validate this hypothesis, we evaluated models trained with \ourmethod{}'s preference data against state-of-the-art LLMs on BigGen Bench, using GPT-4-Turbo, GPT-4, and the Prometheus-eval/prometheus-bgb-8x7b-v2.0 model as judge models, with the results shown in Table \ref{tab:tom}.

Our findings reveal that DPO trained with \ourmethod{}'s preference data outperforms all open-source models and performs close to the GPT-4 series in terms of theory of mind capability. Additionally, when trained using the SimPO objective, \ourmethod{} achieves a 7.4\% relative performance gain in this area. Compared to other models based on Mistral-7B, \ourmethod{} also demonstrates strong improvements in theory of mind capability. Overall, by leveraging preference data derived from UGC, \ourmethod{} effectively captures implicit user intentions, beliefs, and emotions, significantly enhancing the model's ability to understand these aspects and improving its theory of mind performance.

% \paragraph{\ourmethod{} performance on Big Gen Bench}
% wildbench

\begin{table}[t]
    \centering
    \caption{Performance of \ourmethod{} across different ablated versions trained with the DPO objective. \textit{Claude-3-Sonnet Gen. Inst.} refers to using Claude-3-Sonnet to generate and filter reader instructions for improved quality. These results validate the effectiveness of our design choices, highlighting \ourmethod{}'s robustness to instruction quality and the crucial role of reward model selection.}
    \resizebox{1\linewidth}{!}{
    \begin{tabular}{l c c c }
        \toprule[1.5pt] 
         \multirow{2}{*}{\textbf{Ablation Settings}} & \multicolumn{3}{c}{\textbf{Alpaca Eval 2.0}}  \\
         \cmidrule(r){2-4}  %\cmidrule(r){6-7} \cmidrule(r){8-9}  
         & LC Win Rate & Win Rate & Avg. Len  \\
        \midrule[0.75pt]
        
       full \ourmethod{} & 35.93 & 33.05 & 1705\\
       \hline
       \textit{w/o} reference UGC & 31.12 & 30.72 & 2072\\
       \textit{w/o} instruction filtering & 28.92 & 28.68 & 2254 \\
       % \textit{w/o filtering} & 32.10 & 34.62 & 2217\\
       \textit{w/} random reference UGC & 16.74 & 16.33 & 2093\\
       Claude-3-sonnet Gen. Inst. & 35.71 & 34.00 & 1845 \\
       % reward model
       Skywork-Llama-8B as RM + \textit{w/} UGC & 16.12 & 17.08 & 2396 \\
       Skywork-Llama-8B as RM + \textit{w/o} UGC & 14.75 & 15.04  & 3437 \\
       % ArmoRM-Llama3-8B as RM & 5.86 & 6.59 & 4402\\
       \bottomrule[1.5pt] 
    \end{tabular}
    }
    
    \label{tab:ablation}
\end{table}

\paragraph{Ablation Study}
Given \ourmethod{}'s strong performance in alignment benchmarks, we conduct an ablation study to assess the impact of key design choices. As shown in Table \ref{tab:ablation}, removing UGC as references results in a large performance drop, highlighting the importance of the implicit user preference signals in UGC. Removing the instruction relevance filtering step or using randomly selected UGC as references also result in a notable performance drop, highlighting the alignment between instruction and corresponding implicit user preferences in UGC.
To investigate the impact of instruction quality, we utilize a strong proprietary model, Claude-3-Sonnet, for generating and filtering instructions based on UGC. The model trained with Claude-3-Sonnet-generated instructions performs similarly to the default setup using SFT baseline-generated instructions, indicating that instruction quality has a limited impact on alignment performance, consistent with observations of \citet{ivison2024unpacking}. 
For reward model ablation, we replace {Prometheus-7b-v2.0} with {Skywork-Reward-Llama-3.1-8B}, a top RM on RewardBench \cite{lambert2024rewardbench}, and present results with and without reference UGC. Despite its strong RewardBench performance, Skywork struggles with UGC-extracted prompts and encourages generating overly long responses with low win rates, raising concerns about its generalization to unseen prompts. Adding UGC as references moderates response length and improves win rates, demonstrating the value of UGC as implicit preference signals. However, Skywork still underperforms Prometheus, likely because Prometheus explicitly incorporates reference answers during training, allowing it to better evaluate responses relative to references, highlighting the necessity of reference-based judge training in \ourmethod{}.
% For the reward model ablation, we replace {prometheus-7b-v2.0} with {Skywork-Reward-Llama-3.1-8B}, a leading classifier reward model on RewardBench \cite{lambert2024rewardbench}. Despite strong results on RewardBench, Skywork underperforms within \ourmethod{}, where preference data generated with both models produce excessively long responses and low win rates. This highlights the need for a stronger reward model capable of utilizing reference text input, which is critical for enhancing the performance of \ourmethod{}.

\section{Online Iterative Training Can Further Improve Alignment Performance}
While our main experiments focus on the offline setting, where all UGC is used for a single iteration update, online iterative RLHF has demonstrated stronger performance compared to the offline approach \cite{dong2024rlhf, wu2024meta, yuan2024self}. In this section, we present the results of the online iterative training process to evaluate the effectiveness of our proposed \ourmethod{} in this setting. Specifically, we followed the instruction generation and filtering process from the main experiment, using the same 60k instructions and corresponding UGC. The data was randomly split into three subsets, each used for one iteration of training. After each iteration, we sampled responses from the updated model for reward scoring. As shown in Table \ref{tab:iterative}, each iteration yielded steady improvements in LC win rates, with gains of 13.74\%, 3.22\%, and 3.44\% for each respective iteration. Compared to the 35.93\% LC win rate in the offline setting, \ourmethod{} achieved a 37.51\% win rate by the third iteration. Additionally, we observed a significant increase in generation length in the online iterative setting, indicating the need for a better length control method.

\begin{table}[t]
    \centering
    \caption{\ourmethod{}'s performance on Alpaca Eval 2.0 benchmark under online iterative training setting. We find the online iterative setting can further improve the model performance armed with \ourmethod{}.}
    \resizebox{1\linewidth}{!}{
    \begin{tabular}{l c c c }
        \toprule[1.5pt] 
         \multirow{2}{*}{\textbf{Ablation Settings}} & \multicolumn{3}{c}{\textbf{Alpaca Eval 2.0}}  \\
         \cmidrule(r){2-4}  %\cmidrule(r){6-7} \cmidrule(r){8-9}  
         & LC Win Rate & Win Rate & Avg. Len  \\
        \midrule[0.75pt]
        SFT & 17.11 & 14.72 & 1676\\
        \hline
       Offline (1 Iteration) & 35.93 & 33.05 & 1705\\
       \hline
       \ourmethod{}+Iter1 & 30.85 & 28.62 & 1821 \\
        \ourmethod{}+Iter2 & 34.07 & 35.48 & 2200 \\
        \ourmethod{}+Iter3 & 37.51 & 42.31 & 2587 \\
       \bottomrule[1.5pt] 
    \end{tabular}
    }
    
    \label{tab:iterative}
\end{table}

\begin{table}[t]
    \centering
    \caption{Performance of Mistral-7B-Instruct-v0.2 SFT model, on-policy UltraFeedback alignment baseline, and our \ourmethod{} method on SafetyBench \cite{zhang-etal-2024-safetybench} under zero-shot English setting.}
    \resizebox{1\linewidth}{!}{
    \begin{tabular}{l c c c c c c c | c}
        \toprule[1.5pt] 
         \textbf{Model} & \textbf{EM} & \textbf{IA} & \textbf{MH} &  \textbf{OFF} &  \textbf{PH} &  \textbf{PP} &  \textbf{UB}  & \textbf{Avg.}\\
        \midrule[0.75pt]
      SFT & 68.4& 74.3 & 77.4 & 69.1 & 72.7 & 75.8 & 68.2 & 71.9 \\
      UltraFeedback & 67.4 & 72.6 & 78.2 & 70.6 & 69.6 & 76.8 & 61.0 & 70.4 \\ 
      {\ourmethod{}+Unsafe UGC} & 67.9 & 73.5 & 77.0 & 68.9 & 71.8 & 75.3 & 67.4 & 71.5 \\
      \hline
      \textbf{\ourmethod{}} & 68.4 & 74.2 & 77.5 & 69.5 & 72.5 & 75.9 & 68.2 & 72.0 \\
       \bottomrule[1.5pt] 
    \end{tabular}
    }
    \label{tab:safetybench}
\end{table}

\section{Safety Performance of \ourmethod{}}
\label{sec:safety}
To assess whether user-generated content poses potential safety and toxicity issues, we use the comprehensive safety evaluation benchmark SafetyBench \cite{zhang-etal-2024-safetybench}. We evaluate the performance of three models: the Mistral-7B-Instruct base model, the model aligned using UltraFeedback on-policy preference data, and our model aligned using \ourmethod{} generated preference data. The results in Table \ref{tab:safetybench} indicate that, compared to the SFT base model, the model aligned using UltraFeedback data performs worse in terms of safety, whereas the safety performance of our \ourmethod{} model remains largely unchanged with a slight improvement. Moreover, we further include some contaminated data from \citet{ji2024beavertails} into user-generated content to test the robustness of \ourmethod{} against harmful user-generated content. Specifically, we use the responses in the qa-classification task, randomly downsample 1,000 unsafe responses, and add them to the user-generated content for \ourmethod{} pipeline. The corresponding model's performance is denoted as {\ourmethod{}+unsafe UGC} in Table \ref{tab:safetybench}, which shows slight safety degradation compared to \ourmethod{} trained on pure Dolma corpus, yet performs better than model trained with UltraFeedback on-policy preference data. 
These results suggest that although the large UGC corpus may be noisy and may contain toxic or unsafe content, our \ourmethod{} method is robust to such issues, achieving safety performance that is similar to or slightly better than the base model and even safer than the traditional on-policy UltraFeedback alignment approach. We speculate that the unsafe outputs are primarily due to unsafe content in the LLM pretraining data rather than in the alignment data \cite{lin2023unlocking}.

\section{Related Work}
\paragraph{Reinforcement Learning from Human Feedback}
RLHF is a technique that aligns LLMs with human values \cite{christiano2017deep, ziegler2019fine, bai2022training}. The classical RLHF pipeline typically comprises three stages: supervised finetuning \cite{alpaca, kopf2024openassistant, wang2023openchat}, reward model training \cite{gao2023scaling, lightman2023let, lambert2024rewardbench}, and policy model optimization \cite{schulman2017proximal, anthony2017thinking}. Proximal Policy Optimization (PPO) \cite{schulman2017proximal} is a widely used algorithm for policy model optimization and is applied to mitigate toxicity \cite{korbak2023pretraining}, ensure safety \cite{dai2023safe}, enhance factuality \cite{tian2023fine} and reasoning capabilities \cite{havrilla2024teaching}. Given that online preference optimization algorithms are complex and difficult to optimize \cite{zheng2023secrets, santacroce2023efficient}.
Researchers proposed Direct Preference Optimization (DPO) \cite{rafailov2024direct} to avoid training the reward model, and instead directly train the LLM using human preferences. A variety of preference optimization objectives have been proposed besides DPO, including RRHF \cite{yuan2024rrhf}, SLiC-HF \cite{zhao2023slic}, IPO \cite{azar2024general}, CPO \cite{xu2024contrastive}, KTO \cite{ethayarajh2024kto}, ORPO \cite{hong2024reference}, R-DPO \cite{park2024disentangling}, SimPO \cite{meng2024simpo}, and WPO \cite{zhou2024wpo}.

\paragraph{Alignment Data Generation and Curation}
% The alignment data can be categorized into supervised finetuning (SFT) data for instruction following and preference data for human value alignment. 
% % IFT data
Alignment data can be broadly classified into two categories: supervised fine-tuning (SFT) data for instruction following and preference data for aligning models with human values. SFT data typically comes from human-annotated sources like Open Assistant \cite{kopf2024openassistant}, Super-NI \cite{wang2022super}, FLAN \cite{longpre2023flan}, and Natural Instructions \cite{mishra2021cross}, though large-scale manual annotation is costly. To address this, researchers generate synthetic instructions \cite{wang2023self, xu2023wizardlm, mitra2023orca}, while BackHump \cite{li2023self} creates SFT data by back-translating unlabeled web corpora. \citet{nguyen2024better} improves response diversity and quality with rewriting techniques.
% . However, manually annotating data at scale is expensive, prompting researchers to synthesize large quantities of instructions \cite{wang2023self, xu2023wizardlm, mitra2023orca}. BackHump \cite{li2023self} introduces unlabeled web corpora and back-translates text into instructions for SFT data generation. \citet{nguyen2024better} further enhances response diversity and quality through rewriting processes.
% preference data
For preference data, researchers have experimented with augmenting datasets using trained SFT policies \cite{zhao2023slic} or refined SFT policies coupled with rejection sampling \cite{liu2023statistical}, allowing the policy to learn from data generated by optimal policies. Some studies reveal that offline DPO and PPO methods cannot strategically explore the environment \cite{xiong2023gibbs}, emphasizing the value of suboptimal, on-policy data \cite{tajwar2024preference, yu2024lions}. As a result, preference optimization has been extended into iterative online settings, continuously updating reference models with the latest policy models \cite{kim2024sdpo, dong2024rlhf, xiong2024iterative, yuan2024self}. Sourced from user data, WildFeedBack \cite{shi2024wildfeedback} generates preference data from real-time, in-situ user interaction.

Building on insights from UGC that reflect human values and preferences \cite{tan2023user, tan2024democratizing}, we are the first to construct preference data from unlabeled UGC, offering a cost-effective method to generate high-quality data using implicit user preferences.
% Previous work reveal that user-generated content contains insightful information and reflects human intentions, values, and beliefs, making it a good source for automated LLM alignment with human preferences \cite{tan2023user, tan2024democratizing}. In this work, we are the first to construct preference data from unlabeled user-generated content, facilitating a more cost-efficient way to construct high-quality preference data leveraging the implicit user preference signals in the UGC.

\section{Conclusion}
We introduced \ourmethod{}, a framework that leverages implicit user preferences from UGC to generate high-quality preference data for aligning LLMs with human values. By transforming UGC into reader queries and sharer-preferred responses, \ourmethod{} enhances preference alignment. It achieved a state-of-the-art 35.93\% length-controlled win rate on Alpaca Eval 2.0 based on Mistral-7B-Instruct, and consistently outperformed models trained with traditional preference data. Additional studies demonstrated improvements in reward quality, UGC robustness, domain-specific flexibility, and theory of mind capabilities. Our proposed framework opens new opportunities for leveraging implicit human preferences from unlabeled text to align LLMs more effectively with human values.

\section*{Acknowledgement}
This work was supported by an internship at Amazon.com Inc and NSF IIS-2119531, IIS-2137396, IIS-2142827, IIS-2234058, CCF-1901059, and ONR N00014-22-1-2507.

\section*{Limitations}
We identify two key limitations in \ourmethod{}. First, while it is designed to optimize LLM generation quality and align outputs with human preferences using UGC, UGC may contain toxic, harmful, or unsafe content. \ourmethod{} does not explicitly address safety or honesty, which are critical in real-world applications. Future work should focus on integrating safety and honesty considerations into the alignment process when leveraging UGC for preference learning.
Second, \ourmethod{} doesn't show improvement in math, coding, and reasoning tasks. As shown in Tables \ref{tab:bgb_all} and \ref{tab:more_benchmarks}, \ourmethod{} underperforms compared to the SFT model on GSM8k and Coding\&Debugging tasks from WildBench \cite{lin2024wildbench}. We hypothesize two possible reasons for this: \textit{i)} the scarcity of high-quality UGC that provides strong preference signals in reasoning-intensive domains like coding and math, and \textit{ii)} the limitations of the reward model, Prometheus-7b-v2.0, which was trained primarily to evaluate chat and generation quality, rather than reasoning accuracy \cite{kim2023prometheus}. Developing a more robust reward model with better generalization across domains, particularly one that can handle reasoning tasks while utilizing reference text, remains an area for future exploration. Our work mainly focuses on aligning LLMs to understand users' implicit preferences in UGC and enhance the theory of mind capabilities. Given \ourmethod{}’s demonstrated effectiveness in domain-specific alignment, we argue that with access to suitable user-generated content in reasoning-intensive domains and a reward model optimized for reasoning, \ourmethod{} has the potential to synthesize high-quality data that could significantly improve LLM reasoning performance.

\section*{Ethical Considerations}
\paragraph{Data Quality and Bias} Leveraging UGC to align LLMs with human preferences introduces the risk of amplifying existing biases present in the data. UGC, often sourced from diverse and unmoderated platforms, may contain biased, harmful, or misleading content. If these biases are not identified and mitigated, they could be reflected in the model’s outputs, potentially leading to unfair or prejudiced responses. Ensuring that the UGC used for preference alignment is representative, diverse, and free of harmful biases is crucial to prevent perpetuating negative effects. Future work should focus on bias detection and mitigation strategies when using UGC for training.

\paragraph{Safety and Misinformation} UGC can sometimes include harmful, toxic, or incorrect information, which poses significant challenges when aligning LLMs with such content. Although \ourmethod{} enhances alignment by extracting implicit preferences from UGC, it does not explicitly account for the safety, accuracy, or truthfulness of the information being processed. This could inadvertently lead to the generation of unsafe or misleading responses. It is worth noting that the primary goal of our work is to present a proof of concept for leveraging implicit user preferences in user-generated content to align LLMs with user preferences. It is not designed to specifically address the challenges posed by harmful, toxic, or biased UGC. Our focus is on demonstrating the potential of user-generated content as a valuable resource for alignment, rather than solving the separate, complex problem of mitigating harmful content. Harmful, toxic, and biased content constitutes only a small fraction of overall user-generated content, leaving an abundant and valuable resource for alignment. For example, during the first half of 2024, X (Twitter)'s global transparency report for H1 2024\footnote{\url{https://transparency.x.com/content/dam/transparency-twitter/2024/x-global-transparency-report-h1.pdf}} revealed that only 0.0123\% of posts violated platform rules over six months. Similarly, research on Reddit has shown that approximately 16\% of users publish toxic posts and 13\% users post toxic comments \cite{almerekhi2022investigating}. These statistics indicate that harmful UGC forms a minority, and our focus on relatively clean user-generated content does not diminish our contribution of showcasing its value for aligning LLMs with user preferences.
Moreover, we include an additional study in Appendix \ref{sec:safety} to study the safety performance of including harmful content in UGC corpus, results indicate that model trained with our \ourmethod{} is safer than trained with traditional UltraFeedback on-policy preference data. Though adding harmful content can degrade the safety performance of \ourmethod{}, it is still better than the UltraFeedback baseline, indicating the robustness of \ourmethod{} against harmful content.
Future improvements could incorporate explicit mechanisms to ensure that aligned models not only reflect human preferences but also prioritize safety, truthfulness, and the exclusion of harmful content.

\paragraph{Privacy and Consent} UGC often originates from individuals who may not be aware that their content is being used to train AI models. This raises important privacy and consent issues, especially when personal or sensitive information is involved. Even when anonymized, user-generated content may carry implicit private information that could be exposed through model outputs. Ensuring that UGC is collected and used in a way that respects privacy and obtains appropriate consent from content creators is essential. Strong privacy safeguards should be implemented to protect users' data throughout the alignment process.

\bibliography{custom}
\clearpage
\appendix

\section{Reward Score Distribution}
\label{app:reward_distribution}

\begin{figure}[t]
    \centering
    \includegraphics[width=1\linewidth]{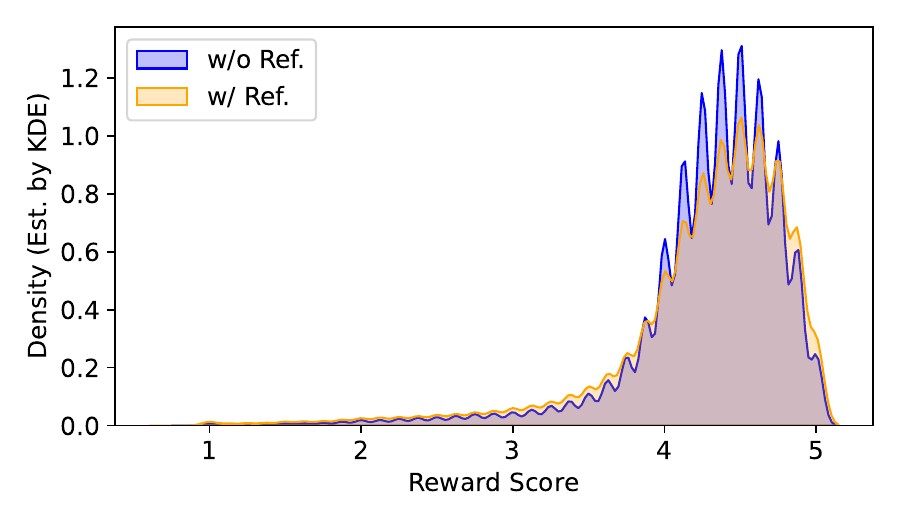}
    \caption{We compare the reward score distribution with and without UGC as a reference. We find that using UGC as a reference results in a smoother reward score distribution, providing a more fine-grained reward signal.}
    \label{fig:reward_distribution}
\end{figure}

To further evaluate the reward quality, we visualize the reward distribution for both cases, with and without UGC as reference text, using prompts generated from UGC, as shown in Figure \ref{fig:reward_distribution}. We observe that the reward distribution without UGC reference is sharper, while the distribution with UGC reference is smoother. This suggests that incorporating implicit user preference signals from UGC facilitates more fine-grained reward signals.

\begin{table}[t]
    \centering
    \caption{Agreement with human preferences under different self-consistency sampling numbers and with/without user-generated content as references under both with-tie and without-tie settings.}
    \resizebox{1\linewidth}{!}{
    \begin{tabular}{c c c c c c}
        \toprule[1.5pt] 
         \multicolumn{2}{c}{\textbf{Settings}} & \multicolumn{4}{c}{\textbf{Self-Consistency ($N$)}}  \\
         \cmidrule(r){1-2}
         \cmidrule(r){3-6}  %\cmidrule(r){6-7} \cmidrule(r){8-9}  
        w/ Tie & w/ UGC & 1 & 3 & 5 & 8  \\
        \midrule[0.75pt]
        & & 0.298 & 0.429 & 0.497 & 0.571 \\
        & $\checkmark$ & 0.393 & 0.497 & 0.592 & 0.649 \\ 
        $\checkmark$ & & 0.550 & 0.571 & 0.584 & 0.634 \\
        $\checkmark$ & $\checkmark$ & 0.594 & 0.626 & 0.665 & 0.691\\
       \bottomrule[1.5pt] 
    \end{tabular}
    }
    
    \label{tab:human_agreement}
\end{table}

\section{Agreement with Human Annotators}

To further validate the quality of the reward model and assess the impact of user-generated content, we conducted a human study to examine the reward quality agreement with actual human preferences using the same data as in our GPT-4-Turbo agreement experiment in Section \ref{sec:reward_quality}. In this study, two graduate students annotated their preferences for 400 data pairs. The raw annotations showed a 0.725 agreement between human preferences and GPT-4 annotations, demonstrating reasonable alignment and validating our experimental design choice. To ensure high-quality human annotations, we retained only the data points where both annotators agreed, resulting in 191 data points.

Building on this, we analyzed the agreement between human preferences and reward given by reward model (Prometheus-2-7B) under varying self-consistency sampling numbers (N) to show the reward quality, as shown in Table \ref{tab:human_agreement}. The results demonstrate that under both with and without tie settings, human agreement of reward signals using user-generated content as references consistently outperformed those without user-generated content. These findings further validate the improvement introduced by leveraging implicit preferences from user-generated content and its alignment with genuine human preferences. 

\begin{figure}
    \centering
    \includegraphics[width=1\linewidth]{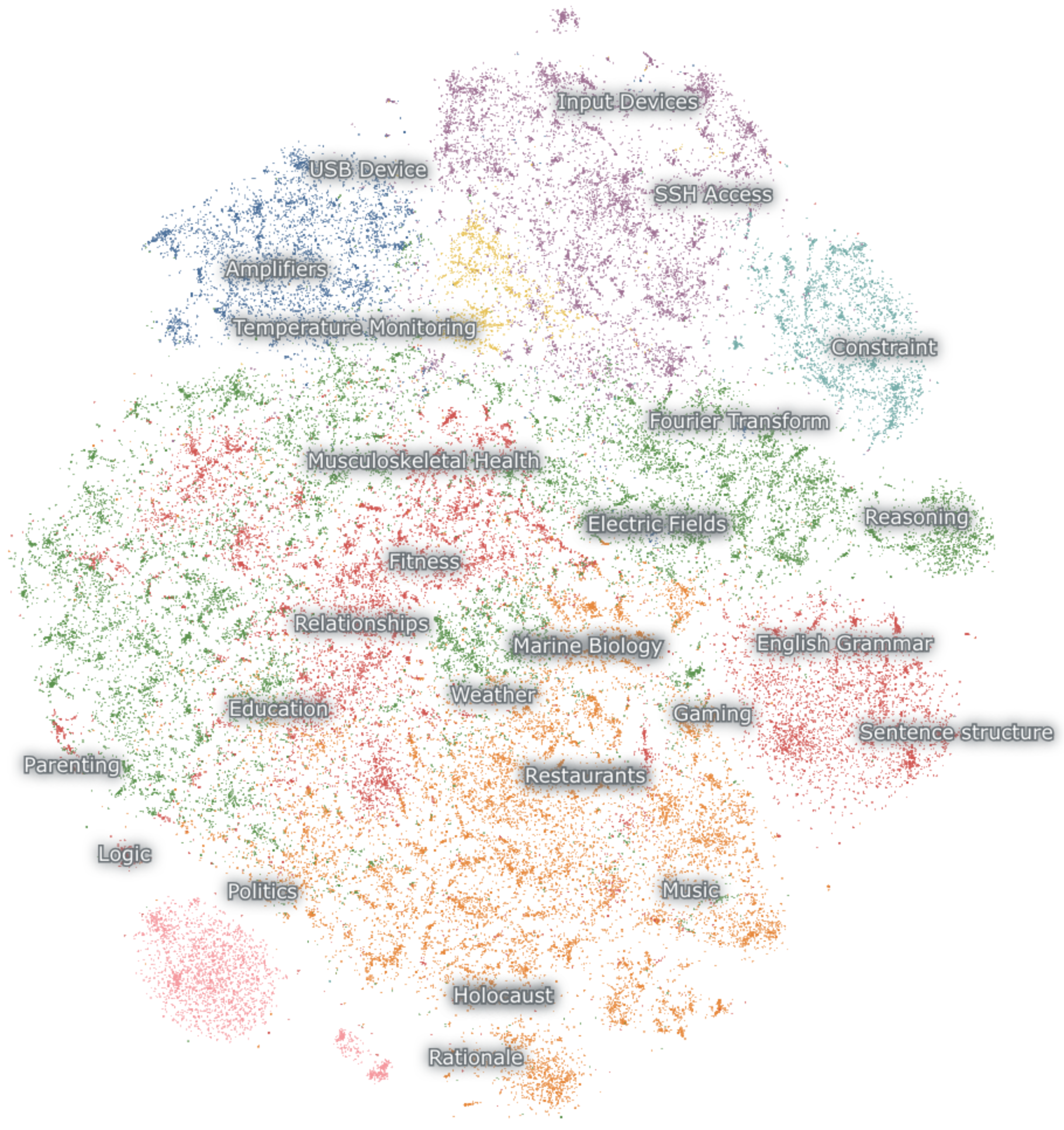}
    \caption{Visualization of our collected UGC data via Nomic Atlas. The figure is colored by topics, which are auto-generated by the Nomic Atlas custom topic model.}
    \label{fig:ugc_vis}
\end{figure}

% TBD
% To validate the potential safety and toxicity issues lie in the user-generated content, we further employ a comprehensive safety evaluation benchmark, SafetyBench \cite{zhang-etal-2024-safetybench}, to test the performance of Mistral-7B-Instruct base model, model that aligned using UltraFeedback on-policy preference data, and our \ourmethod{} generated preference data. Results presented in Table \ref{tab:safetybench} show that compared to SFT base model, model that aligned using UltraFeedback prompt shows worse safety performance, while our \ourmethod{} safety performance remain relatively the same and shows slight improvement. These results show that though the massive UGC corpus can be noisy and potentially contains toxic and unsafe content, our \ourmethod{} is robust to potentially unsafe content and shows the same and even slight improvement to the base model, it's even safer to the traditional on-policy UltraFeedback alignment paradigm. We speculate that the unsafe generation was mainly sourced from the unsafe content in LLM pretraining data, rather than the alignment data.

\begin{table}[t]
    \centering
    \caption{AlpacaEval 2.0 results for different reward models and prompt settings.}
    \resizebox{1\linewidth}{!}{
    \begin{tabular}{l l c c c c}
        \toprule[1.5pt]
        \textbf{RM} & \textbf{Prompt} & \textbf{UGC Ref.} & \textbf{LC} & \textbf{WR} & \textbf{Len.} \\
        \midrule[0.75pt]
        Prometheus & UGC Gen. & \Checkmark  & 35.93 & 33.05 & 1705 \\
        Prometheus & UGC Gen. & \XSolidBrush & 31.12 & 30.72 & 2072 \\
        Prometheus & UltraFeedback & \XSolidBrush & 26.56 & 24.39 & 1750 \\
        Skywork & UGC Gen. & \Checkmark & 16.12 & 17.08 & 2396 \\
        Skywork & UGC Gen. & \XSolidBrush & 14.75 & 15.04 & 3437 \\
        Skywork & UltraFeedback & \XSolidBrush & 30.85 & 28.62 & 1821 \\
        \bottomrule[1.5pt]
    \end{tabular}
    }
    \label{tab:RM}
\end{table}

\section{Reward Model Analysis}
Our proposed \ourmethod{} framework aims to leverage the implicit user preference signals in the user-generated content to \textit{enhance} LLM alignment with human preferences, but not entirely independent of direct human supervision. To further investigate the role of the reward model, we include additional studies in Table \ref{tab:RM}. Specifically, we evaluated prompts derived from UGC in our \ourmethod{} framework but performed reward scoring using Skywork without UGC as references. Additionally, we examined the performance of using Skywork as a reward model and scoring responses sampled from UltraFeedback prompts. From these results, we have the following observations:

\paragraph{For reward model choice, Prometheus outperforms Skywork} Prometheus outperforms Skywork across all settings. Since Prometheus explicitly incorporates references during training, these results highlight the necessity of reference-based reward model training in our PUGC framework.
\paragraph{Skywork struggles with UGC-derived prompts} Despite its strong RewardBench performance, Skywork performs poorly on UGC-derived prompts, leading to excessively long responses with low win rates. This raises concerns about its generalization capability to unseen/out-of-distribution prompts. These results also demonstrate that the unsatisfying performance when using Skywork as a reward model is limited by the generalization of the Skywork model rather than our PUGC framework.
\paragraph{UGC as a reference improves performance} Under both Prometheus and Skywork, incorporating UGC as a reference improves win rates and moderates response length compared to scoring without UGC references. This demonstrates the value of UGC as an implicit preference signal for preference alignment.

Together, these results show that:
\begin{itemize}[leftmargin=*]
\item The poor performance observed when using Skywork as a reward model is \textit{not a limitation of our PUGC framework but rather reflects Skywork’s limited generalization capabilities when applied to UGC-generated prompts}.
\item Incorporating UGC as a reference, as done in PUGC, consistently improves alignment performance and yields more moderate response lengths, underscoring the advantage of leveraging implicit preferences in UGC. Moreover, as demonstrated in the domain-specific experiment in Section \ref{domain_specific_alignment}, \ourmethod{} enables more flexible and domain-specific alignment and enhances the reward model generalization using domain-specific prompts.
\item Prometheus provides strong alignment performance, likely due to its training process, which includes reference answers and helps the model learn the correlation between reference UGC and responses. Given that reference-based evaluation is crucial for robust preference modeling, Prometheus is currently the only existing reward model that meets this criterion. We encourage future research to explore the development of reference-based reward models to further advance alignment techniques.
\end{itemize}

\paragraph{The practicality of \ourmethod{}} We do not claim that \ourmethod{} replaces on-policy UltraFeedback or other preference data. Instead, it provides a flexible way to use unlabeled user-generated content to complement existing methods. Researchers can still train reward models with on-policy data when it is beneficial. In practice, \ourmethod{} can serve as an additional source of preference signals, enabling \textit{more flexible and domain-specific preference alignment}, which is supported by domain-specific UGC alignment performance in Section \ref{domain_specific_alignment}.

\section{Selected UGC Distribution}
We visualize the distribution of the collected UGC using Nomic Atlas\footnote{\url{https://atlas.nomic.ai/}} and present the results in Figure \ref{fig:ugc_vis}. The UGC demonstrates a broad distribution across various domains, with the majority of data concentrated in areas such as technology and devices, health and medical topics, education and language, social and political issues, entertainment and leisure, as well as legal and historical subjects. This distribution highlights the model's significant performance improvements in theory of mind capabilities, enabling a deeper understanding of users' intentions and beliefs. However, the scarcity of reasoning-related data limits the model's effectiveness in reasoning-intensive tasks, such as mathematics and coding.

\begin{table}[t]
    \centering
    \caption{Breakdown of Alpaca Eval test set instructions by instruction category.}
    \resizebox{1\linewidth}{!}{
    \begin{tabular}{l r r}
        \toprule[1.5pt] 
         \textbf{Category} & \textbf{Number} & \textbf{Percentage} \\
        \midrule[0.75pt]
        general knowledge & 55 & 6.8\%\\
        historical/cultural topics & 37 & 4.6\%\\
        explanations/reasoning& 130 & 16.1\%\\
        directions/navigation& 24 & 3.0\%\\
        categorization/classification & 59 & 7.3\%\\
        product/service descriptions & 68 & 8.4\%\\
        scientific concepts& 63 & 7.8\%\\
        writing tasks & 19 & 2.4\%\\
        creative writing & 87 & 10.8\%\\
        math problems & 35 & 4.3\%\\
        review/critique & 17 & 2.1\%\\
        email/letter writing & 18 & 2.2\%\\
        language/grammar & 33 & 4.1\%\\
        hypothetical scenarios & 34 & 4.2\%\\
        programming/coding & 76 & 9.4\$\\
        social media content & 14 & 1.7\%\\
        summarization & 10 & 1.2\% \\        
        game/puzzle solutions & 6 & 7.4\% \\
        script/dialogue writing & 9 & 1.1\% \\
        data analysis/interpretation & 9 & 1.1\% \\
       \bottomrule[1.5pt] 
    \end{tabular}
    }
    \label{tab:alpaca_task}
\end{table}

\begin{table}[t]
    \centering
    \caption{Breakdown of Alpaca Eval test set instructions by instruction complexity.}
    \resizebox{0.6\linewidth}{!}{
    \begin{tabular}{l r r}
        \toprule[1.5pt] 
         \textbf{Category} & \textbf{Number} & \textbf{Percentage} \\
        \midrule[0.75pt]
        1 & 2 & 0.2\%\\
        2 & 17 & 2.1\%\\
        3 & 72 & 8.9\%\\
        4 & 95 & 11.8\%\\
        5 & 59 & 7.3\%\\
        6 & 214 & 26.6\%\\
        7 & 251 & 31.2\%\\
        8 & 92 & 11.4\%\\
        9 & 1 & 0.1\%\\
       \bottomrule[1.5pt] 
    \end{tabular}
    }
    \label{tab:alpaca_complexity}
\end{table}

\begin{table}[t]
    \centering
    \caption{Breakdown of Alpaca Eval test set instructions by expected response length.}
    \resizebox{0.9\linewidth}{!}{
    \begin{tabular}{l r r}
        \toprule[1.5pt] 
         \textbf{Category} & \textbf{Number} & \textbf{Percentage} \\
        \midrule[0.75pt]
        1 sentence &23 & 2.9\%\\
        1-3 sentences& 177 & 22.0\%\\
        1 paragraph& 322 & 40.0\%\\
        2 paragraphs& 252 & 31.3\%\\
        3 or more paragraphs & 29 &3.6\% \\
       \bottomrule[1.5pt] 
    \end{tabular}
    }
    \label{tab:alpaca_length}
\end{table}

\begin{figure}[t]
    \centering
    \includegraphics[width=1\linewidth]{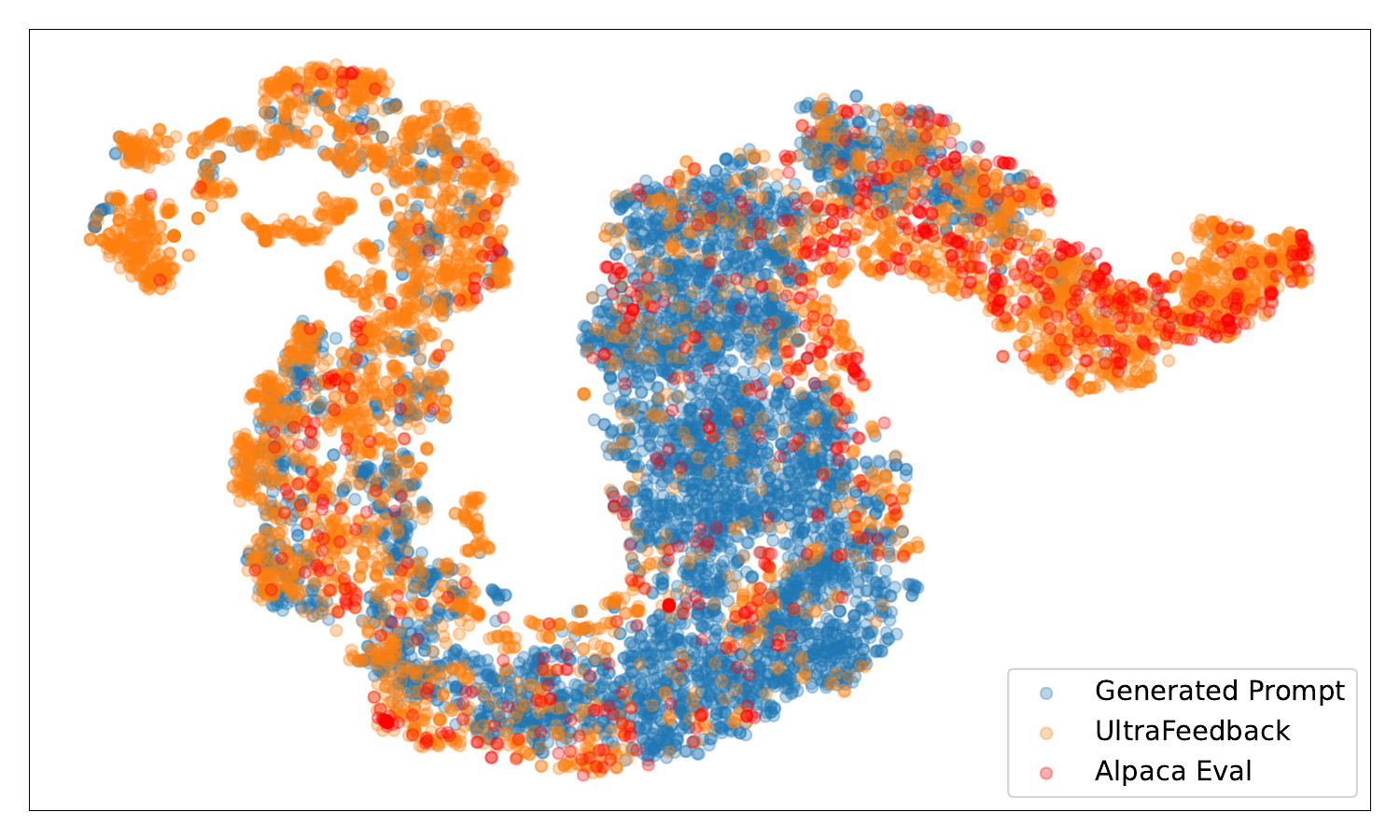}
    \caption{T-SNE visualization of generated prompt from \ourmethod{}, UltraFeedback, and Alpaca Eval. The embedding of the prompts are calculated by DeBERTa-v3-Large \cite{he2021debertav3}. Our generated prompt are closer to the Alpaca Eval test set prompt, while UltraFeedback prompts distribute more diverse.}
    \label{fig:prompt_vis}
\end{figure}

\section{Generated Query Distribution Analysis}
To better understand the distribution of generated query from UGC, and its relation with prompts from UltraFeedback and Alpaca Eval test set, we visualize the prompt distribution in Figure \ref{fig:prompt_vis}. Specifically, we use DeBERTa-v3-Large \cite{he2021debertav3} for prompt encoding and apply t-SNE to visualize the prompt embeddings. Our results show that the prompts generated by \ourmethod{} are more closely aligned with those from the Alpaca Eval test set, while the UltraFeedback prompts exhibit greater diversity. This observation aligns with the findings in \citet{nguyen2024better}.

\section{Fine-grained Alpaca Eval Data Statistics}
We use Claude-3-sonnet to categorize prompt in Alpaca Eval test set based on task category, instruction complexity, and expected response length. The category statistics can be found in Table \ref{tab:alpaca_task}, \ref{tab:alpaca_complexity}, and \ref{tab:alpaca_length}. We discard categories that contain less than 10 samples to ensure the evaluation quality.

\begin{table*}[t]
    \centering
    \caption{Performance of \ourmethod{} on WildBench \cite{lin2024wildbench} and Arena-Hard \cite{arenahard2024}.}
    \resizebox{1\linewidth}{!}{
    \begin{tabular}{l c c c c c c c}
        \toprule[1.5pt] 
         \multirow{2}{*}{\textbf{Model}} & \multicolumn{6}{c}{\textbf{WildBench}} & \multirow{2}{*}{\textbf{Arena-Hard}} \\
         \cmidrule(r){2-7}    
         & WB score & Creative Tasks & Planning\&Reasoning & Math\&Data Analysis & Information/Advice Seeking & Coding\&Debugging  \\
        \midrule[0.75pt]
        Mistral-7B-instruct & 25.63 & 42.07 & 30.06 & 10.08 & 40.10 & 18.4 & 12.6\\
        % UltraFeeback+DPO & \\
        \ourmethod{}+DPO & 26.95 & 46.56 & 33.36 & 11.43 & 40.20 & 17.16 & 13.7\\
       
       \bottomrule[1.5pt] 
    \end{tabular}
    }
    \label{tab:wildbench_arenahard}
\end{table*}

\begin{table*}[t]
    \centering
    \caption{Performance of \ourmethod{} on BigGen Bench \cite{kim2024biggen} evaluate by prometheus-bgb-8x7b-v2.0.}
    \resizebox{1\linewidth}{!}{
    \begin{tabular}{l c c c c c c c c c c}
        \toprule[1.5pt] 
       \textbf{Model} & \textbf{Ground.} & \textbf{Inst. Follow.} & \textbf{Plan.} &\textbf{Reason.} & \textbf{Refine.} & \textbf{Safe.} & \textbf{ToM} & \textbf{Tool.} & \textbf{Multi.} & \textbf{Overall}\\
        \midrule[0.75pt]
        Mistral-7B-instruct & 3.73 & 3.84 & 3.63 & 3.11 & 3.26 & 3.75 & 3.66 & 3.21 & 2.33 & 3.40\\
        UltraFeedback+DPO & 3.64 & 3.68 & 3.80 & 3.02 & 3.29 & 3.22 & 3.79 & 2.49 & 2.49 & 3.27 \\
       \ourmethod{} + DPO & 3.71 & 3.90 & 3.87 & 3.09 & 3.29 & 3.20 & 4.30 & 3.09 & 2.39 & 3.43\\
       \bottomrule[1.5pt] 
    \end{tabular}
    }
    \label{tab:bgb_all}
\end{table*}

% \section{Training Details}
% We 
\section{Decoding Hyperparameters}
\label{decode_hyper}
In preference data generation, we set temperature to 0.8 with top\_p to 0.95 under the base setting and temperature to 0.7 and top\_p to 0.9 for readers' instruction generation based on UGC. For instruction filter, we restrict the model's output token to \{True, False\} and do greedy decoding with max\_token=1. For reward score generation, we follow the default hyperparameters in Prometheus \cite{kim2024prometheus} and set temperature to 1.0 with top\_p=0.9.

In response generation for evaluation, we follow the setting in SimPO \cite{meng2024simpo}. Specifically, for Alpaca Eval 2.0, we use a sampling decoding strategy to generate responses, with a temperature of 0.7 for the Mistral-Base setting following zephyr-7b-beta, and a temperature of 0.5 for the Mistral-Instruct setting following Snorkel-Mistral-PairRM-DPO. For MT-Bench, we follow the official decoding configuration which defines different sampling temperatures for different categories.

% \section{\ourmethod{} Performance Across Diverse Benchmarks}

\begin{table}[t]
    \centering
    \caption{Performance of \ourmethod{} on GSM8k \cite{cobbe2021training}, MMLU-Pro \cite{wang2024mmlu}, IFEval \cite{zhou2023instruction}, and TruthfulQA \cite{lin2021truthfulqa}.}
    \resizebox{1\linewidth}{!}{
    \begin{tabular}{l c c c c c c }
        \toprule[1.5pt] 
         \textbf{Model} & \textbf{GSM8k} & \textbf{MMLU-Pro} & \textbf{IFEval} & \textbf{TruthfulQA}\\
        \midrule[0.75pt]
        Mistral-7B-instruct & 43.06 & 30.42 & 38.83 & 52.26\\
        % UltraFeedback+DPO & \\
        \ourmethod{}+DPO & 41.17 & 28.37 & 40.11 & 42.77 \\
       
       \bottomrule[1.5pt] 
    \end{tabular}
    }
    \label{tab:more_benchmarks}
\end{table}

\section{Computation Resources Details}
All the training experiments in this paper were conducted on a single node with 8 × NVIDIA A100-SXM4-40GB GPUs.

\section{Scientific Artifacts}
\ourmethod{} is built with the help of many existing scientific artifacts, including PyTorch \cite{paszke2019pytorch}, Numpy \cite{harris2020array}, huggingface, and transformers \cite{wolf2020transformers}, alignment-handbook \cite{Tunstall_The_Alignment_Handbook}, and RLHFlow \cite{dong2024rlhf}. We will make the \ourmethod{} implementation publicly available to facilitate further research.

\section{Performance on Additional Benchmarks}
We also evaluate \ourmethod{} on additional benchmarks, including WildBench \cite{lin2024wildbench}, Arena-Hard \cite{arenahard2024}, BigGen Bench \cite{kim2024biggen}, GSM8k \cite{cobbe2021training}, MMLU-Pro \cite{wang2024mmlu}, IFEval \cite{zhou2023instruction}, and TruthfulQA \cite{lin2021truthfulqa}. The results are presented in Tables \ref{tab:wildbench_arenahard}, \ref{tab:bgb_all}, and \ref{tab:more_benchmarks}. 

In WildBench, \ourmethod{} demonstrates significant improvements in creative tasks and reasoning \& planning, while showing comparable performance in math \& data analysis and information/advice seeking. We hypothesize that \ourmethod{} focuses on leveraging implicit user preferences from UGC for alignment, rather than enhancing coding and data analysis capabilities. A similar trend is observed in the Arena-Hard benchmark, which primarily consists of math and coding instructions. 

The BigGen Bench results in Table \ref{tab:bgb_all} indicate stable performance in grounding, instruction following, planning, reasoning, refinement, and multilingual tasks. However, we observe a notable gain in theory of mind capabilities, supporting the hypothesis that UGC provides rich information about implicit human beliefs, values, and preferences. On the other hand, we observe performance drops in safety and tool use, which are not the focus of this work, and leave improvements in these areas for future research.

For GSM8k, MMLU-Pro, IFEval, and TruthfulQA, shown in Table \ref{tab:more_benchmarks}, we note a performance drop in GSM8k, MMLU-Pro, and TruthfulQA, potentially due to the alignment tax \cite{lin2023speciality, ouyang2022training}. However, moderate improvement is observed in IFEval, indicating stronger instruction-following capabilities in \ourmethod{}.

\begin{table}[t]
\centering
\caption{Correlation Matrix and Summation Across Different Language Models}
\resizebox{1\linewidth}{!}{
\begin{tabular}{l c c c c c}
\toprule[1.5pt]
\textbf{Model} & \textbf{Llama} & \textbf{Mistral} & \textbf{Mixtral} & \textbf{Claude-3} & \textbf{Corr. Sum}\\
\midrule[0.75pt]
\textsc{Llama-3-70B-it} & - & 0.31 & 0.31 & 0.49 & 1.11\\
\textsc{Mistral-Large} & 0.31 & - & 0.25 & 0.46 & 1.02\\
\textsc{Mixtral-8x7B-it} & 0.31 & 0.25 & - & 0.27 & 0.84\\
\textsc{Claude-3-Sonnet} & 0.49 & 0.46 & 0.27 & - & 1.23\\
\bottomrule[1.5pt]
\end{tabular}
}
\label{tab:model_correlations}
\end{table}

\begin{table*}[t]
    \centering
    \caption{AlpacaEval 2.0 Performance of using \textsc{Mistral-Nemo-Instruct-2407} as policy model under DPO tuning.}
    \resizebox{1\linewidth}{!}{
    \begin{tabular}{lccc}
        \toprule[1.5pt] 
        \textbf{Model} & \textbf{Win Rate (\%)} & \textbf{LC Win Rate (\%)} & \textbf{Avg. Length} \\
        \midrule[0.75pt]
        {Mistral-Nemo-Instruct-2407} & 41.34 & 44.64 & 1887 \\
        {UltraFeedback + Mistral-Nemo-Instruct-2407} & 57.86 & 52.81 & 2546 \\
        {\ourmethod{} + Mistral-Nemo-Instruct-2407 (Ours)} & 60.95 & 54.99 & 2414 \\
        \bottomrule[1.5pt] 
    \end{tabular}
    }
    
    \label{tab:policy_model_results}
\end{table*}

\section{UGC Quality Assessment Model Selection}
\label{sec:ugc_quality_selection}
For the UGC quality assessment model, since there is no ground truth for quality ratings, we evaluated the correlation between the quality scores generated by different strong LLMs to identify the most representative filtering model. Specifically, we randomly sampled 1,000 user-generated content instances and asked these LLMs to assign quality scores using the same prompt. We then calculated pairwise correlation between their ratings.

From the correlation present in Table \ref{tab:model_correlations}, we find that Claude-3-Sonnet is the most representative model. However, due to its proprietary nature and high cost for large-scale quality filtering, we chose the second-best model, Llama-3-70B, which is open-source and can be locally deployed. Llama-3-70B offers an optimal balance between cost and performance for our use case.

\begin{table}[t]
    \centering
    \caption{The agreement of UGC quality assessment with human annotations using different judge models. We find the chosen LLama-3-70B-Instruct shows moderate agreement with human annotation and significantly outperform Mixtral-8x7B-Instruct.}
    \resizebox{1\linewidth}{!}{
    \begin{tabular}{lccc}
        \toprule[1.5pt] 
    \textbf{Agreement w/ Human} & \textbf{Accuracy} & \textbf{Cohen's Kappa} \\
    \midrule[0.75pt] 
    {Mixtral-8x7B-Instruct} & 0.54 & 0.08 \\ 
    {Llama-3-70B-Instruct} & 0.74 & 0.48 \\ \bottomrule[1.5pt]
    \end{tabular}
    }
    \label{tab:UGC_quality_valid}
\end{table}

\section{Validation for UGC Quality Assessment}

\label{sec:ugc_quality_validation}

To further validate the effectiveness of our quality filtering methodology, we manually annotate the quality of UGC and present the UGC quality assessment agreement humans and the selected LLM. Specifically, we sample 100 data pairs with a quality score difference greater than 2 and ask human annotators to perform pairwise judgments, identifying which UGC better reflects implicit human preferences.
The agreement between \texttt{Llama-3-70B} scoring and human judgment in Table \ref{tab:UGC_quality_valid} shows that \texttt{Llama-3-70B} demonstrates moderate agreement with humans.  
The selected annotation model, \texttt{Llama-3-70B}, also shows a stronger correlation with human annotators compared to \texttt{Mixtral-8x7B-Instruct-v0.1}, further validating the effectiveness of our UGC quality filtering method. We will include these results in the final version to validate the high-quality claim.

\section{Performance with Additional Policy Models}

To validate the generalization of \ourmethod{} with different policy models, we also experimented with using \texttt{mistralai/Mistral-Nemo-Instruct-2407} as a policy model in \ourmethod{} and evaluated its performance under DPO tuning. Results in Table \ref{tab:policy_model_results} demonstrate that our proposed \ourmethod{} framework outperforms on-policy UltraFeedback preference data when using \texttt{Mistral-Nemo-Instruct-2407} as the policy model, further validating the generalization capability of \ourmethod{} across different models.

% \begin{table}[t]
%     \centering
%     \caption{Performance of \ourmethod{} on GSM8k \cite{cobbe2021training}, MMLU-Pro \cite{wang2024mmlu}, IFEval \cite{zhou2023instruction}, and TruthfulQA \cite{lin2021truthfulqa}.}
%     \resizebox{1\linewidth}{!}{
%     \begin{tabular}{l c c c c c c }
%         \toprule[1.5pt] 
%          \textbf{Model} & \textbf{GSM8k} & \textbf{MMLU-Pro} & \textbf{IFEval} & \textbf{TruthfulQA}\\
%         \midrule[0.75pt]
%         Mistral-7B-instruct & 43.06 & 30.42 & 38.83 & 52.26\\
%         % UltraFeedback+DPO & \\
%         \ourmethod{}+DPO & 41.17 & 28.37 & 40.11 & 42.77 \\
       
%        \bottomrule[1.5pt] 
%     \end{tabular}
%     }
%     \label{tab:more_benchmarks}
% \end{table}

\section{Case Study}
We present examples of preference data generated from UGC sourced from Dolma in Tables \ref{tab:dolma_ugc1}, \ref{tab:dolma_ugc2}, and \ref{tab:dolma_ugc3}, as well as Pref data derived from Goodreads book reviews in Tables \ref{tab:goodreads_ugc1}, \ref{tab:goodreads_ugc2}, and \ref{tab:goodreads_ugc3}. 

In Dolma's preference data, the chosen response highlights implicit human preferences in UGC by focusing on trends like digital subscriptions, social media, and video content, which reflect user demands for convenience, personalization, and engagement. The emphasis on AI, machine learning, and voice assistants indicates a growing preference for automation, while the resurgence of local news shows a desire for community-focused reporting. These insights capture the shift toward diverse, adaptive, and immersive content experiences.

In Goodreads' preference data, the response mirrors insightful human preferences by emphasizing detailed evaluations of novels, such as language richness, character complexity, and thematic depth. It reflects readers’ appreciation for vivid imagery and emotional engagement, while also considering accessibility and discussion potential, aligning with UGC’s focus on immersive, thought-provoking content.

\newpage

\begin{prompt}{Prompt for UGC Quality Scoring}
Below is a text document. Evaluate whether the document is a good source for identifying potential user queries and crafting corresponding responses to demonstrate how an AI Assistant should handle user instructions. Please assign a score using the following 5-point scale:\\

1: The document is incomplete, vague, off-topic, or does not address what the user might want to know. It lacks any originality or valuable insights, making it difficult to extract relevant queries or responses. For example, the content might be missing key information, be poorly organized, or contain irrelevant or promotional text.\\

2: The document has some potential but does not provide all the necessary information to extract relevant queries or responses. It might offer a high-level overview or general ideas but lacks depth or specificity. \\

3: The document is helpful and covers basic topics but lacks depth or insightful information. It addresses common user concerns but does so in a generic or superficial manner. While it is self-contained and complete, it may not offer substantial content for crafting nuanced AI Assistant responses. \\

4: The document is a strong source for extracting potential user queries and crafting responses. It provides a clear, comprehensive, and well-organized presentation of information. While the content is informative and useful, there may still be minor areas for improvement, such as clarity, conciseness or focus.\\

5: The document is a perfect source for extracting potential user queries and responses. It is high-quality, demonstrating expert knowledge and offering clear, logical, and engaging insights. The document is infused with creative and original ideas that challenge conventional thinking, making it a perfect model for how an AI Assistant should respond to user instructions. \\

Please first provide a brief reasoning you used to derive the rating score, and then write "Score: " on the last line.\\

<document start>\\
\{\texttt{User-Generated Content}\}\\
<document end>
\end{prompt}

% \section{Prompt for Reader Question Generation}
% \label{app:gen_q}

\begin{prompt}{Prompt for Reader Question Generation}
\label{sec:gen_q}

\#\#\#Task Instruction\\
Act as a human and generate a question or instruction based on the provided context. Here is an instruction or query example, you can follow its pattern.\\ 

<example start> \{\texttt{Instruction Example}\} <example end>\\

<context start> \{\texttt{User-Generated Content}\} <context end>\\

Ensure that the provided context contains sufficient information and details to support responses to the question or instruction. Your question or instruction should be complete, independent, and self-contained. Give specific and detailed information in the question or instruction. Only generate one question or instruction and nothing else. The question or instruction cannot explicitly refer to the context.
    
\end{prompt}

\begin{prompt}{Prompt for Question Filtering}
\#\#\#Context\\
\{\texttt{User-Generated Content}\}\\

\#\#\#Question\\
\{\texttt{Generated Instruction}\}\\

\#\#\#Task Instruction\\
Determine if the context contains accurate, comprehensive, profound, creative, and insightful information relevant to the question. Respond with only True or False without any additional characters or formatting.

\end{prompt}

\begin{prompt}{GPT Judge Prompt Template}
<|im\_start|>system\\
You are a highly efficient assistant, who evaluates and selects the best large language model (LLMs) based on the quality of their responses to a given instruction. This process will be used to create a leaderboard reflecting the most accurate and human-preferred answers.\\
<|im\_end|>\\
<|im\_start|>user\\
I require a leaderboard for various large language models. I'll provide you with prompts given to these models and their corresponding outputs. Your task is to assess these responses, and select the model that produces the best output from a human perspective.\\

\#\# Instruction\\

\{\\
    \verb|    |``instruction'': ````\texttt{\{instruction\}}'''',\\
\}\\
\\

\#\# Model Outputs\\

Here are the unordered outputs from the models. Each output is associated with a specific model, identified by a unique model identifier.\\

\{\\
    \verb|    |\{\\
    \verb|    |\verb|    |    ``model\_identifier'': ``m'',\\
    \verb|    |\verb|    |    ``output'': ````{\texttt{\{Output\_1\}}}''''\\
\verb|    |\},\\
\verb|    |\{\\
    \verb|    |\verb|    |    ``model\_identifier'': ``M'',\\
    \verb|    |\verb|    |    ``output'': ````{\texttt{\{Output\_2\}}}''''\\
\verb|    |\}\\
\}\\
\\

\#\# Task\\

Evaluate the models based on the quality and relevance of their outputs, and select the model that generated the best output. Answer by providing the model identifier of the best model. We will use your output as the name of the best model, so make sure your output only contains one of the following model identifiers and nothing else (no quotes, no spaces, no new lines, ...): m or M.\\

\#\# Best Model Identifier\\
<|im\_end|>
\end{prompt}

\begin{prompt}{Reward Scoring Prompt Template}
\#\#\#Task Description:\\
An instruction (might include an Input inside it), a response to evaluate, a reference document that contains relevant information, and a score rubric representing a evaluation criteria are given.\\
1. Write a detailed feedback that assess the quality of the response strictly based on the given score rubric, not evaluating in general.\\
2. After writing a feedback, write a score that is an integer between 1 and 5. You should refer to the score rubric.\\
3. The output format should look as follows: "(write a feedback for criteria) [RESULT] (an integer number between 1 and 5)"\\
4. Please do not generate any other opening, closing, and explanations.\\

\#\#\#The instruction to evaluate:\\
\texttt{\{Instruction\}}\\

\#\#\#Response to evaluate:\\
\texttt{\{Response\}}\\

\#\#\#Reference Document:\\
\texttt{\{Reference\_UGC\}}\\

\#\#\#Score Rubrics:\\
\texttt{\{Rubric\}}\\

\#\#\#Feedback:
\end{prompt}

\begin{prompt}{Rubric Template}
[\texttt{\{criteria\}}]\\
Score 1: \texttt{\{score1\_description\}}\\
Score 2: \texttt{\{score2\_description\}}\\
Score 3: \texttt{\{score3\_description\}}\\
Score 4: \texttt{\{score4\_description\}}\\
Score 5: \texttt{\{score5\_description\}}
\end{prompt}

\begin{prompt}{Score Rubric}
\texttt{criteria}:"Does the response meet the criteria of quality, considering factors such as helpfulness, relevance, accuracy, depth, creativity, and level of detail?"\\

  \texttt{score1\_description}: "The response is incomplete, vague, off-topic, or not aligned with the user's request. It may contain missing content, start lists incorrectly, or repeat the user's question. The response might reflect another person’s perspective, include irrelevant information such as promotional text, or resemble a forum post rather than a direct answer."\\
  
  \texttt{score2\_description}: "The response addresses most of the user’s requests but does not directly answer the question. It may provide a high-level methodology instead of the exact solution, leaving the user without the specific guidance they were seeking."\\
  
  \texttt{score3\_description}: "The response is useful but appears to be written by someone other than an AI assistant. It meets the basic requirements of the user's question but is written from another person’s perspective, resembling content from a blog post, web page, or search results rather than an AI-generated answer." \\
  
  \texttt{score4\_description}: "The response is crafted from the perspective of an AI assistant, with a clear focus on addressing the user’s instruction. It provides a complete, clear, and comprehensive answer without missing or irrelevant information. While well-organized and helpful, there may be minor areas for improvement in clarity, conciseness, or focus."\\
  
  \texttt{score5\_description}: "The response is an exemplary AI-generated answer, perfectly aligned with the user’s instruction. It is focused, helpful, and free of irrelevant content, delivering high-quality insights and expert knowledge. The response is engaging, logical, insightful, and easy to follow, reflecting the highest standard of quality for an AI assistant."
\end{prompt}

\begin{prompt}{Prompt used to obtain instruction categories in the AlpacaEval test set.}
\label{app:alpaca_finegrained}
\texttt{<LIST ALL ALPACAEVAL INSTRUCTIONS>}\\
Given the above list of possible instructions, define a maximum of 20 categories that would cover the types of instructions, for example recipes, reasoning tasks, general knowledge etc.\\
Try to cover as many of the instructions as possible with the maximum 20 categories, while keeping the categories high-level, simple and easy to understand.
\end{prompt}

\begin{prompt}{Prompt for categorizing instructions.}

Instruction: \texttt{\{Instruction\}}\\

Given the above, categorize it into one of the following 20 categories:\\

\texttt{<LIST ALL CATEGORIES>}\\

Secondly, score the instruction in terms of complexity: how complex you think it
is to answer from 1-10 (where 10 is a complex question whereby first reasoning or breaking
down the question into multiple subquestions for example might help improve the answer).\\

Thirdly, indicate how long you think the response to the instruction should be, either (a) 1 sentence, (b) 1-3 sentences, (c) 1 paragraph, (d) 2 paragraphs, or (e) 3 or more
paragraphs.\\

Provide your final response in the following format:\\
Category: <one of the 20 categories>\\
Complexity: <score out of 10>\\
Length: <length category>. Do not provide the actual response.
    
\end{prompt}

\begin{prompt}{Prompt for Domain-Specific Evaluation}

[System]\\
Please act as an impartial judge and evaluate the quality of the responses provided by two AI assistants to the user question displayed below. You should choose the assistant that follows the user's instructions and answers the user's question better. Your evaluation should consider factors such as the helpfulness, relevance, accuracy, depth, creativity, and level of detail of their responses. Begin your evaluation by comparing the two responses and provide a short explanation. Avoid any position biases and ensure that the order in which the responses were presented does not influence your decision. Do not allow the length of the responses to influence your evaluation. Do not favor certain names of the assistants. Be as objective as possible. After providing your explanation, output your final verdict by strictly following this format: "[[A]]" if assistant A is better, "[[B]]" if assistant B is better, and "[[C]]" for a tie.\\

[User Question]\\
\texttt{\{Question\}}\\

[The Start of Assistant A's Answer]

\texttt{\{Answer\_a\}}

[The End of Assistant A's Answer]\\

[The Start of Assistant B's Answer]

\texttt{\{Answer\_b\}}

[The End of Assistant B's Answer]

\end{prompt}

\begin{small}

\begin{table*}[t]
\caption{Preference data example from Dolma UGC.}
\begin{adjustbox}{max width=1\linewidth}
\begin{tabular}{p{1.5in}p{5in}}
    \toprule[1.5pt]
    \textbf{User-Generated Content} &  We all wish we had the power to predict what the world will look like in five, 10 or 50 years.
    As a sector currently undergoing turbulent change, the media industry is frantically shaking a cloud-filled crystal ball up and down in the hopes of figuring out what the future holds.
    The afternoon-long workshop provided us with a glimpse of how the IFTF goes about the forecasting process. It turns out that future-gazing is not as mystical as one might think.
    The goal of the workshop – to explore the future of news and journalism – was largely focused on brainstorming. Together with staff and affiliate researchers from the IFTF, we took it in turns to share “signals” for the future of the media.
    A signal is an existing example that indicates the shape of the future. e.g. a prototype service, a lab demo or a current news item.
    The task wasn’t simply to share ideas about the latest media-centric iPhone app or crowd-sourced content strategy. Marina and her team also charged us with inferring what an existing signal says about a future trend for journalism.
    Over the course of about 90 minutes, we came up with a voluminous slew of signals and trends, filling the long white-board at the front of IFTF’s workspace with dozens of colorful sticky notes. Marina and her colleague Mike Liebhold (a distinguished fellow at the Institute) then tried to organize the ideas into clusters representing various trend areas.
    The breadth of ideas we came up with suggests nothing if not an excitingly multi-dimensional future for our industry – a future that is already in embryonic stage today.
    For example, the existence of norran.se, a Swedish website which enables readers to participate in the editorial decision-making process, and the “sousveillance” movement, point to deeper public involvement in news-gathering and disseminating in the years ahead; the demise of the personal computer and rise of the “deckchair-friendly” tablet interface suggest a bigger market for the consumption of longer-form journalism; meanwhile, the increasing prevalence of non-native-language news sources in various parts of the world illustrates a trend towards the trans-border news interests of a switched-on society that is deeply curious about how people in other cultures feel and think.
    A robust future?
    From capitalizing on new interfaces and platforms for the dissemination of news to creating innovative forms of algorithmic journalism that tailor information to suit a user’s location, interests and even mood, the media industry has plenty of room for a robust future.
    Of course, while there is no shortage of interesting ways in which our sector can grow in the years ahead, one thing remains cloudy in that crystal ball: how anyone is going to make any money. Not even the augurs at IFTF have a handle on that.\\
    \hline 
    \textbf{Generated Instruction}& What trends in the media industry, indicated by specific signals, suggest new ways for news-gathering, dissemination, and consumption, and how might these trends impact the revenue model for media companies?\\
    \hline
    \textbf{Chosen Response} &  1. Digital Subscriptions: The trend of consumers paying for digital content is on the rise. According to a report by PwC, global digital subscription revenue in the media industry is expected to reach \$52 billion by 2024. This shift towards digital subscriptions is a significant change from the traditional revenue model of advertising-driven revenue.
    \newline
    2. Social Media: Social media platforms have become an integral part of news-gathering, dissemination, and consumption. According to a study by Pew Research Center, 62\% of US adults get news on social media. This trend has significant implications for media companies as they need to adapt to this new distribution channel and find ways to monetize it. \newline
    3. Video Content: Video content is becoming increasingly important in the media industry. According to a report by Cisco, video will account for 82\% of all internet traffic by 2022. Media companies are investing heavily in video production and distribution to meet the growing demand for video content.\newline
     \texttt{(continued)}
     \\
    % \hline
    % \textbf{Rejected Response} & \\    
     \bottomrule[1.5pt]
\end{tabular}
\end{adjustbox}
\label{tab:dolma_ugc1}
\end{table*}

\begin{table*}[t]
\caption{Preference data example from Dolma UGC (continued).}
\begin{adjustbox}{max width=1\linewidth}
\begin{tabular}{p{1.5in}p{5in}}
    \toprule[1.5pt]
    \textbf{Chosen Response} \texttt{(continued)} &\texttt{(continued)} 4. Podcasts: Podcasts are gaining popularity as a new medium for news and entertainment. According to a report by Edison Research, 51\% of Americans have listened to a podcast, up from 40\% in 2018.
    Media companies are exploring new revenue streams through sponsorships and subscriptions for podcasts. 5. Artificial Intelligence and Machine Learning: AI and ML are being used to personalize news content for individual consumers. This trend is expected to continue as media companies look for ways to differentiate themselves in a crowded market. However, the use of AI and ML also raises ethical concerns around privacy and data security.\newline 6. Voice Assistants: Voice assistants such as Amazon Alexa and Google Assistant are becoming increasingly popular for news consumption. According to a report by eMarketer, 24.5 million US adults will use a voice assistant for news in 2022. Media companies are exploring new ways to deliver news through voice assistants and monetize this new distribution channel.\newline 7. Local News: Local news is experiencing a resurgence as consumers look for more personalized and community-focused content. According to a report by Pew Research Center, 43\% of US adults get local news daily. Media companies are investing in local news reporting and delivery to meet this demand.\newline Overall, these trends suggest a shift towards personalized and subscription-based news consumption, with a focus on digital and video content. Media companies need to adapt to these trends by investing in new technologies, delivery channels, and business models to remain competitive. However, these trends also come with challenges around privacy, data security, and ethical considerations that need to be addressed.\\ 
    \hline
    \textbf{Rejected Response} & 1. Digital Subscriptions: The shift towards digital consumption of news and the success of subscription models, as seen in the case of The New York Times, The Washington Post, and The Guardian, suggest a new revenue model for media companies. These subscriptions provide a steady stream of revenue, unlike traditional advertising models that rely on high traffic and impressions.\newline
    2. Social Media as News Sources: The increasing use of social media as a news source indicates a trend towards decentralized news-gathering and dissemination. This trend challenges traditional media companies' monopoly on news production and distribution. It also creates new opportunities for revenue generation through targeted advertising and partnerships with social media platforms.\newline
    3. Podcasting: The rise of podcasting as a popular form of news consumption indicates a trend towards audio content. This trend provides an opportunity for media companies to explore new revenue streams through podcast advertising and sponsorships.\newline 
    4. Artificial Intelligence and Automated Journalism: The use of artificial intelligence (AI) and automated journalism indicates a trend towards data-driven journalism and content generation. This trend can lead to cost savings for media companies and increased efficiency, but it also raises ethical concerns regarding accuracy and fact-checking.\newline
    5. Video Content: The trend towards video content, as seen in the success of YouTube and streaming services, suggests a shift towards visual storytelling and entertainment. This trend creates new opportunities for media companies to generate revenue through advertising and subscription models. However, it also requires significant investment in production and distribution infrastructure.\newline
    6. Personalization: The trend towards personalized content and recommendations, as seen in the success of Netflix and Spotify, suggests a future where media consumption is tailored to individual preferences. This trend provides opportunities for media companies to generate revenue through targeted advertising and subscription models based on user data.\newline
     \texttt{(continued)}\\    
     \bottomrule[1.5pt]
\end{tabular}
\end{adjustbox}
\label{tab:dolma_ugc2}
\end{table*}

\begin{table*}[t]
\caption{Preference data example from Dolma UGC (continued).}
\begin{adjustbox}{max width=1\linewidth}
\begin{tabular}{p{1.5in}p{5in}}
    \toprule[1.5pt]
    \textbf{Rejected Response} \texttt{(continued)} &
    \texttt{(continued)} 7. Localized News: The trend towards localized news and hyperlocal journalism indicates a shift towards community-focused reporting and away from national and international news. This trend creates new opportunities for revenue generation through targeted advertising and subscription models based on local demographics and interests.\newline
    8. Programmatic Advertising: The trend towards programmatic advertising, which uses algorithms to target specific audiences, suggests a future where advertising is more efficient and effective. This trend provides opportunities for media companies to generate revenue through targeted advertising based on user data.\newline 9. Virtual and Augmented Reality: The trend towards virtual and augmented reality content suggests a future where news and entertainment are more immersive and interactive. This trend creates new opportunities for revenue generation through advertising and subscription models based on user engagement and experience.\newline
    10. Blockchain Technology: The trend towards blockchain technology in the media industry suggests a future where content is verified, secure, and transparent. This trend creates new opportunities for revenue generation through subscription models based on trust and security. It also provides opportunities for new business models based on microtransactions and tokenization. \\    
     \bottomrule[1.5pt]
\end{tabular}
\end{adjustbox}
\label{tab:dolma_ugc3}
\end{table*}
\end{small}

\begin{small}
    \begin{table*}[t]
\caption{Preference data example from Goodreads UGC.}
\begin{adjustbox}{max width=1\linewidth}
\begin{tabular}{p{1.5in}p{5in}}
    \toprule[1.5pt]

    \textbf{User-Generated Content} & SECOND READING: Sept 2, 2014 \newline
    I've just re-read THE GOLDFINCH by Donna Tartt because I chose this book for my book club to discuss and I wanted to refresh my memory on its details. The first time I read it (see my original review below), I loved it for many reasons. While most of them still hold true for me, I have to say I didn't find it worthy of the 2014 Pulitzer Prize for Fiction which it won since my previous review. \newline
 Donna Tartt is a storyteller extraordinaire. I love the depth of her details and the reaches of her imagination. I am surprised however by her abundant usage of adverbs and cliches. This alone should have knocked it out of contention for the Pulitzer, in my opinion. Beyond that, while I grew extremely tired of Theo and Boris's use of drugs during my first reading, I accepted it because I was so into the story. The second time through (plus hearing how tired everyone else grew of it too) made me realize that it was truly way over done. \newline
 I would recommend this book only to people who love an epic novel. I would also knock this down to 4.5 Stars if Goodreads would let me. \newline
 FIRST READING: Jan 31, 2014 \newline
 You know how sometimes a person, or a thought, or a book hits your life at just the right moment, and you learn something from him/her/it? And you somehow know that if it had happened at any other time in your life, it might never have touched you in the same way? Well, THE GOLDFINCH hit me at just the right moment. I loved this book. \newline
 THE GOLDFINCH is 771 pages. Is that too long? Probably. Yet I can't think of a single scene I would remove. Even during the passages where I wondered, What's the point of this? Where is it going? Yeah, Donna Tartt always answered those questions in due time. \newline
 Where to begin? The prose, the plot, the characters, the theme? All wonderful and unique. Tartt's description of each character is so vivid, so fantastically specific that I could picture each one in turn. What a lesson in flawed characters! Even the ones we're "supposed" to like, we don't always. And isn't that the way it is in real life? Don't you even dislike your best friend every now and then? Get a little mad at your spouse on occasion? Exactly. \newline
 THE GOLDFINCH is the story of Theo Decker, a thirteen-year-old boy who finds himself in possession of a masterwork of art from the Metropolitan Museum after a terrorist bombing that killed his mother. The way Tartt leads him from that shell-shocked boy to a damaged man of twenty-eight feels so realistic, so deep and probable that if someone told me this was based on a true story, I would probably believe it. Not that I have much in common with Theo or his childhood friend, Boris, or his mentor, Dobie, or any of the other characters (at least not the big stuff), but I loved them all because of how this story ended. The ending was perfect. Plausible, yet not sweetly Hollywood-slick. Though, the book is so descriptive, it could easily be turned into a movie.\newline
 I would love to discuss the theme of this book and the character development (Theo is the perfect example of a reluctant hero), but I don't want to give too much away. If you're debating on this book due to the length, let me say, Yes, it's long and winding, but the journey is fabulous and the ending will feel like a reward. I have a few knocks against the book, but they're not important enough to alter my rating. \newline
 I will be choosing this book for my book club because I am eager to discuss so many elements. Also, if anyone is curious, not only did I read the book, I listened to it on audible.com. The narrator, David Pittu, does an excellent job with a multitude of characters and foreign accents. The audio version added so much to my reading experience. I highly recommend it. \newline
 Five drunk, gritty, gorgeous Stars (Read the book; you'll understand.)\\
 
     \bottomrule[1.5pt]
\end{tabular}
\end{adjustbox}
\label{tab:goodreads_ugc1}
\end{table*}

    \begin{table*}[t]
\caption{Preference data example from Goodreads UGC (continued).}
\begin{adjustbox}{max width=1\linewidth}
\begin{tabular}{p{1.5in}p{5in}}
    \toprule[1.5pt]
    
 \textbf{Generated Instruction} & How would you evaluate the use of descriptive language and character development in a lengthy novel, and what elements would you consider before recommending it to a book club?\\
    \hline
     \textbf{Chosen Response} & Evaluating the use of descriptive language and character development in a lengthy novel involves a close reading and analysis of the text. Here are some elements to consider:\newline 1. Descriptive Language: \newline - Richness and Vividness: Does the author use vivid and rich language to paint a picture of the setting, characters, and actions? Is the language engaging and evocative, making the reader feel immersed in the story? \newline - Consistency: Is the descriptive language consistent throughout the novel? Does it help to establish a tone and mood that is maintained throughout the story?\newline - Sensory Details: Does the author use sensory details effectively to help the reader experience the story through the five senses? Are there compelling descriptions of sights, sounds, smells, tastes, and textures?\newline - Imagery: Does the author use powerful and memorable imagery to help the reader visualize the story? Are there metaphors, similes, and other figurative language that add depth and meaning to the narrative?\newline 
     2. Character Development:\newline- Complexity: Are the characters multi-dimensional and complex, with well-defined motivations, strengths, weaknesses, and backstories? Do they grow and change throughout the novel, or do they remain static?\newline- Consistency: Is the character development consistent with the narrative and the characters' actions and dialogue? Do the characters behave in ways that are true to their nature and the story?\newline- Relatability: Are the characters relatable and sympathetic, or are they distant and unengaging? Do readers care about what happens to them and why?\newline- Interactions: How do the characters interact with each other? Are there compelling relationships and conflicts between them?\newline 
     Before recommending a lengthy novel to a book club, here are some additional elements to consider:\newline - Themes: What are the major themes of the novel, and how do they resonate with readers? Are they timely and relevant, and do they spark meaningful discussion?\newline- Writing Style: Is the writing style accessible and engaging, or is it dense and challenging? Will all members of the book club be able to engage with the text?\newline - Diversity: Does the novel offer diverse perspectives and representations of people and cultures? Does it challenge readers to broaden their understanding of the world?\newline - Accessibility: Is the novel accessible to all members of the book club, or does it require a significant investment of time and resources (e.g., specialized knowledge, research, or translation)?\\
     \hline
    \textbf{Rejected Response} & Evaluating the use of descriptive language and character development in a lengthy novel involves a close reading and analysis of the text. Here are some key elements to consider:\newline 1. Descriptive Language:\newline - Vivid Imagery: Does the author use descriptive language to paint a clear and vivid picture of the settings, characters, and emotions? Is the imagery evocative and transporting, allowing readers to immerse themselves in the story? \texttt{(continued)}\\
     
     \bottomrule[1.5pt]
\end{tabular}
\end{adjustbox}
\label{tab:goodreads_ugc2}
\end{table*}

    \begin{table*}[t]
\caption{Preference data example from Goodreads UGC (continued).}
\begin{adjustbox}{max width=1\linewidth}
\begin{tabular}{p{1.5in}p{5in}}
    \toprule[1.5pt]
    \textbf{Rejected Response} \texttt{(continued)} & \texttt{(continued)} - Discussion Questions: Are there thought-provoking discussion questions that will help the book club explore the novel's themes and characters in depth? Will these questions encourage lively and engaging conversation?\newline
    - Sensory Details: Does the author use all five senses to describe the world of the novel? Are there rich and specific details about sights, sounds, smells, tastes, and textures that bring the story to life?\newline - Figurative Language: Does the author use figurative language effectively to add depth and meaning to the descriptions? Are there metaphors, similes, or other figurative devices that enhance the reader's understanding of the story?\newline
    2. Character Development:\newline - Complexity: Are the characters well-developed and complex? Do they have distinct personalities, motivations, and backstories that make them feel real and relatable?\newline - Consistency: Are the characters consistent in their actions, thoughts, and dialogue throughout the novel? Do they grow and change in believable ways, or do they remain stagnant?\newline - Interactions: How do the characters interact with each other? Are there compelling and dynamic relationships that drive the plot forward and deepen the readers' understanding of the characters?\newline Before recommending a lengthy novel to a book club, here are some additional elements to consider:\newline - Thematic Depth: Does the novel explore complex themes that would lead to rich and engaging discussions? Are there thought-provoking ideas that would resonate with the members of the book club?\newline - Accessibility: Is the novel accessible to a wide range of readers? Is the language complex or dense, making it challenging for some members to engage with?\newline - Diversity: Does the novel offer diverse perspectives and representations? Are there characters from different cultural backgrounds, or does the novel explore issues of race, gender, sexuality, or disability in meaningful ways?\newline - Engaging Plot: Does the novel have a compelling and engaging plot that would keep the members interested and invested in the story?\newline Overall, a lengthy novel that uses rich descriptive language and develops complex and compelling characters, while exploring complex themes and offering diverse perspectives, would make for an excellent book club selection.\\
     
     \bottomrule[1.5pt]
\end{tabular}
\end{adjustbox}
\label{tab:goodreads_ugc3}
\end{table*}
\end{small}

\end{document}